\documentclass[sigconf]{acmart}

\AtBeginDocument{%
  }

\copyrightyear{2024}
\acmYear{2024}
\setcopyright{acmlicensed}
\acmConference[WSDM '24] {Proceedings of the 17th ACM International Conference on Web Search and Data Mining}{March 4--8, 2024}{Mérida, Yucatán, Mexico.}
\acmBooktitle{Proceedings of the 17th ACM International Conference on Web Search and Data Mining (WSDM '24), March 4--8, 2024, Mérida, Yucatán, Mexico}
\acmPrice{15.00}
\acmISBN{979-8-4007-0371-3/24/03}
\acmDOI{10.1145/3616855.3635752}

\usepackage{subfig}
\usepackage{booktabs} 
\usepackage{amsmath}
\usepackage{amsthm}
\usepackage{amsfonts}       
\usepackage{nicefrac}       
\usepackage{xspace}
\usepackage{xcolor}
\usepackage{backnaur}
\usepackage{multirow}
\usepackage{makecell}
\usepackage{appendix}
\usepackage{enumitem}
\usepackage{circledsteps}
\usepackage{color}
\usepackage{colortbl}
\usepackage{graphicx}

\theoremstyle{definition}

\newcommand{\refsec}[1]{\S\ref{#1}} 
\newcommand{\reftab}[1]{Table~\ref{#1}}

\def\eg{\textit{e.g.}\xspace}

\def\etc{\textit{etc.}\xspace}
\def\ie{\textit{i.e.}\xspace}

\setlist[itemize]{noitemsep}
\usepackage{tcolorbox}

\begin{document}


\title{Table Meets LLM: Can Large~Language~Models Understand Structured~Table~Data? A~Benchmark~and~Empirical~Study}



\author{Yuan Sui}
\authornote{Yuan Sui and Mingjie Zhou made their contributions during their internships at Microsoft Research Asia, located in Beijing, China.}
\email{yuan.sui@u.nus.edu}
\affiliation{
    \institution{National University of Singapore}
    \country{Singapore}
}

\author{Mengyu Zhou}
\authornote{Corresponding author.}
\email{mezho@microsoft.com}
\affiliation{
    \institution{Microsoft}
    \country{Beijing, China}
}

\author{Mingjie Zhou}
\authornotemark[1]
\email{mjzhou@connect.hku.hk}
\affiliation{
    \institution{The University of Hong Kong}
    \country{Hong Kong, China}
}

\author{Shi Han}
\email{shihan@microsoft.com}
\affiliation{
    \institution{Microsoft}
    \country{Beijing, China}
}

\author{Dongmei Zhang}
\email{dongmeiz@microsoft.com}
\affiliation{
    \institution{Microsoft}
    \country{Beijing, China}
}

\begin{abstract}
\label{sec:abstract}

Large language models (LLMs) are becoming attractive as few-shot reasoners to solve Natural Language (NL)-related tasks. However, the understanding of their capability to process structured data like tables remains an under-explored area.
While tables can be serialized as input for LLMs, there is a lack of comprehensive studies on whether LLMs genuinely comprehend this data. In this paper, we try to understand this by designing a benchmark to evaluate the structural understanding capabilities of LLMs through seven distinct tasks, \eg, cell lookup, row retrieval and size detection. Specially, we perform a series of evaluations on the recent most advanced LLM models, GPT-3.5 and GPT-4 and observe that performance varied with different input choices, including table input format, content order, role prompting, and partition marks. Drawing from the insights gained through the benchmark evaluations, we propose \textit{self-augmentation} for effective structural prompting, such as critical value / range identification using internal knowledge of LLMs. When combined with carefully chosen input choices, these structural prompting methods lead to promising improvements in LLM performance on a variety of tabular tasks, \eg, TabFact($\uparrow2.31\%$), HybridQA($\uparrow2.13\%$), SQA($\uparrow2.72\%$), Feverous($\uparrow0.84\%$), and ToTTo($\uparrow5.68\%$). We believe that our open source\footnote{Please find code and data of the paper at our temporary repo at \url{https://anonymous.4open.science/r/StructuredLLM-76F3/README.md.} \\\textbf{Please note that we may replace the private preview with an official one in \url{https://github.com/microsoft/TableProvider} at any time.}} benchmark and proposed prompting methods can serve as a simple yet generic selection for future research.
\end{abstract}

\ccsdesc[500]{Information systems~Information retrieval query processing}
\ccsdesc[500]{Computing methodologies~Natural language processing}
\ccsdesc[500]{Computing methodologies~Natural language generation}

\keywords{large language models, semi-structured data, structural understanding capabilities, benchmark}

\maketitle

\section{Introduction}

Structured data consists of plain text blocks organized by predefined structures to compress recurring information. It makes data more manageable and facilitates data analysis and processing by machines. \textbf{Table} is one of such structured data types with many applications such as Table-based Question Answering (TQA)~\citep{dataset_hybridqa, dataset_sqa}, Table-based Fact Verification (TFV)~\citep{dataset_tabfact, model_unifiedSKG}, Table-to-Text~\citep{model_tuta} and Column Type \& Relation Classification~\citep{model_tabbie, model_turl}. The adoption of structured data has significantly contributed to the advancement of information retrieval and knowledge extraction in web mining and content analysis~\cite{web_mining, Trabelsi_2022}.


Prompt engineering has been proven as a highly effective method for in-context learning (ICL). Recent studies, such as ``chain of thoughts'' (CoT)~\citep{model_cot} and ``self-consistency''~\citep{model_self_consistency} or hybrid approaches using both generation and retrieval methods~\citep{aggarwalLetSampleStep2023, zhangPlanningLargeLanguage2023} have demonstrated that LLMs, \eg, GPT-X~\citep{model_gpt3, model_instruct_gpt3} and FlanT5~\citep{model_flan_t5}, can solve complex mathematical reasoning tasks in both zero-shot and few-shot settings. Furthermore, 
\citet{model_table_reasoner} illustrates that by using CoT with LLMs, GPT-3.5 shows impressive performance with just one-shot demonstration on several tabular tasks. These findings have opened new possibilities for the use of LLMs in structured data.


However, previous work has not provided comprehensive studies that examined whether LLMs can truly understand tabular data or given a detailed discussion of the extent to which LLMs have already achieved structural understanding capabilities. Furthermore, despite the remarkable success of LLMs in handling natural languages, their application to tabular data modality presents unique challenges: as different tables define structure and features in distinct ways and often lack straightforward transformation into sequential text (table serialization). 
Based on our survey, we believe that the process of table serialization, along with context and corresponding queries, is highly flexible. That is, there is a lack of grounded consensus or comprehensive investigation on what constitutes a common-sense or exhaustive input design for LLMs on tabular tasks. Previous work used various input designs in an ad-hoc manner~\citep{model_tapas, model_mate, model_table_reasoner, model_tuta, model_tapex, model_tabbie, model_tablegpt}.
For example, TaPEx~\citep{model_tapex} uses special tokens to indicate components like headers <HEAD> and rows <ROW>;
TABBIE~\citep{model_tabbie} serialize tables by both row-wise and column-wise;
while TableGPT~\citep{model_tablegpt} use a template-based method to serialize attribute-value pairs in each table record, \eg, changing ``name: Elon Musk" to ``name is Elon Musk." and concatenating all the sentences according to the order of the records.
The complex landscape of varied input design further complicates the challenges faced by researchers and developers in this field. Therefore, in this paper, our aim is to address the question: \textit{What input designs and choices are the most effective in enabling LLMs to understand tables?}

In this paper, our goal is to address the chaotic landscape of input designs and determine whether LLMs can truly comprehend tabular data. We also aim to discuss the extent to which LLMs have already achieved in terms of their structural understanding capabilities.
To achieve this, we propose a benchmark called \textit{SUC} (structural understanding capabilities) to compare various input designs and create specific tasks in ~\refsec{sec:benchmark} that focus on each structural understanding capability of LLMs. 
To assess the effectiveness of multiple input choices, we conduct a series of experiments using different prompt variants. These variants include input format, format explanation, role prompting, partition mark~\citep{model_uniSAr}, and zero-shot / one-shot approaches. The SUC benchmark offers a comprehensive comparison of multiple input designs, evaluating different aspects of structural understanding capabilities over table(s) as illustrated in \refsec{sec:experiments}. 
We then provide pragmatic guidance on how to better utilize LLM in understanding structured data in ~\refsec{sec:prompting}. Specifically, we propose a \textit{model-agnostic} method called \textit{self-augmented prompting} to directly boost the performance of LLM in downstream tabular-based tasks. This method motivates LLMs to generate intermediate structural knowledge by internally retrieving their own knowledge, \eg, motivates LLMs to generate critical value / range identification by itself. These choices diverge from previous approaches like CoT and Zero-shot-CoT~\citep{model_zero-shot-cot} by focusing on identifying effective methods for unlocking LLMs' capabilities to correctly comprehend structured information.
We find that when combined with carefully chosen input choices, these structural prompting methods lead to promising improvements in LLM performance on various tabular reasoning tasks, \eg, TabFact($\uparrow2.31\%$), HybridQA($\uparrow2.13\%$), SQA($\uparrow2.72\%$), Feverous($\uparrow0.84\%$), and ToTTo($\uparrow5.68\%$) compared to baseline methods. See the results in ~\refsec{sec:experiments}.

Our exploration leads us to believe that 
1) LLMs have basic structural understanding capabilities but are far from perfect, even on trivial tasks, \eg, table size detection (detect the number of columns and rows in a table); 
2) Choosing the right combination of input designs can significantly enhance LLMs' understanding of structured data. Different combinations of serialization functions and input options demonstrate noticeable performance gaps in downstream tasks (see ~\refsec{sec:experiments}). The disparity remains even when using GPT-4, validating the effectiveness of our benchmarking approach; 
3) Self-augmented prompting is a simple model-agnostic method for better utilizing LLMs' internal knowledge and unraveling new possibilities to improve their structural understanding capabilities.
In summary, we propose using markup language like \textbf{HTML} with certain structural features like format explanation and partition mark, combined with self-augmented prompting, to fully leverage LLMs' internal knowledge and achieve better results in tabular reasoning tasks.
\textbf{Our main contributions are:}
\begin{itemize}
    \item We propose the SUC benchmark to evaluate the multiple structural understanding capabilities of LLMs.
    \item Through comprehensive experiments on the benchmark, we provide insights and guidelines on tabular input choices for future work (see ~\refsec{sec:experiments}).
    \item We propose self-augmentation as a method to enhance the performance of LLMs by leveraging internal knowledge. We verify the effectiveness of this simple but generic method on five tabular reasoning datasets.
\end{itemize}

\section{Preliminaries}
\label{sec:preliminaries}

\subsection{Table Structure}

Tabular data exhibit remarkable flexibility in diverse structures, as illustrated in~\citep{survey_table_structure}. These structures include relational tables, entity tables, matrix tables, layout tables, and more. Tables can have horizontal or vertical orientations and span the spectrum from flat to hierarchical. In this paper, we mainly focus on flat relational tables but also have some discussion on hierarchical tables, such as ToTTo~\citep{dataset_totto}. In these tables, each row corresponds to a distinct record, while columns represent specific fields, without any hierarchical arrangement.

Tabular data also exhibit various approaches for formatting values, including text, numbers, date/time, formulas, and other relevant information. In particular, text plays a pivotal role in tables, capturing meta-information such as headers, notes, captions, and cells within the data region. On the other hand, numbers often involve arithmetic relationships like summation and proportion, as well as statistical attributes such as distribution and trends. Furthermore, tables commonly present meticulously organized numerical data, making it easy for reference and comparison. These structured numerical values are often documented using spreadsheet formulas~\citep{survey_tablepretraining}.
The flexibility of tabular data poses unique challenges for LLMs, as different tables define structure and formatting in distinct ways. The gap between tabular data and natural languages (NL) hinders the application of NL reasoning to facilitate table reasoning.

\subsection{Table Serialization \& Splitting}

Table serialization refers to the process of converting data from tables into a linear, sequential text format. This adaptation is essential for training and utilizing LLMs, especially for tasks like masked language modeling, where understanding and predicting language patterns is crucial. A simple serialization function is to serialize tables row-by-row. Many works such as TaPas~\citep{model_tapas}, MATE~\citep{model_mate}, TableFormer~\citep{model_tableformer}, TUTA~\citep{model_tuta}, and TURL~\citep{model_turl} use this method. TaPEx~\citep{model_tapex} uses special tokens to indicate components like headers <HEAD> and rows <ROW>.
TABBIE~\citep{model_tabbie} serialize tables both by row-wise and column-wise. While TableGPT~\citep{model_tablegpt} use a template-based method to serialize attribute-value pairs in each table record.

Furthermore, most LLMs are inefficient in dealing with long sentences due to the quadratic complexity of self-attention~\citep{model_selfattention, survey_effecienttransformer}\footnote{The maximum sequence length of text-Davinci-003 is constrained to 4k tokens.}. However, structured data typically contains dozens of components, which presents a significant challenge in terms of \textit{memory} and \textit{computational efficiency}. While \citet{model_tapex, model_tapas} use naive methods to truncate the input based on a maximum sequence length, this approach may result in the loss of critical information and disrupt the structure of the entire table. In our experiments, we have predefined certain constraints to meet the LLM call request. For example, (1) to avoid potential disruption caused by truncation, we employ a random row sampling strategy when the number of tokens in the table exceeds a certain threshold, and (2) we append a 1-shot example based on the estimated remaining token capacity. Several meticulously crafted sequence serialization functions have been proposed as common practices in table serialization, including \citet{model_tapex, model_tapas, model_tuta, model_formlm, model_unifiedSKG, model_uniSAr}. In this paper, we gather various commonly used serialization methods as baselines and conduct a fair comparison in Sec~\refsec{sec:benchmark}.

\section{SUC Benchmark}
\label{sec:benchmark}

In this section, we aim to develop a benchmark for comparing different input designs and investigating the structural understanding capabilities of LLMs. Specifically, we explore the following aspects: 1) \textit{What input designs and choices are most effective in enabling LLMs to understand tables?}; 2) \textit{To what extent do LLMs already possess structural understanding capabilities for structured data?}
Additionally, we analyze the intricate trade-off of multiple combinations of input designs. Find the benchmark collection and pre-processing details in Sec~\refsec{sec:appendix-suc}.

\subsection{Structural Understanding Capabilities}
\label{sec:capabilities}

\begin{figure}[htbp]
    \centering
    \includegraphics[width=0.9\linewidth]{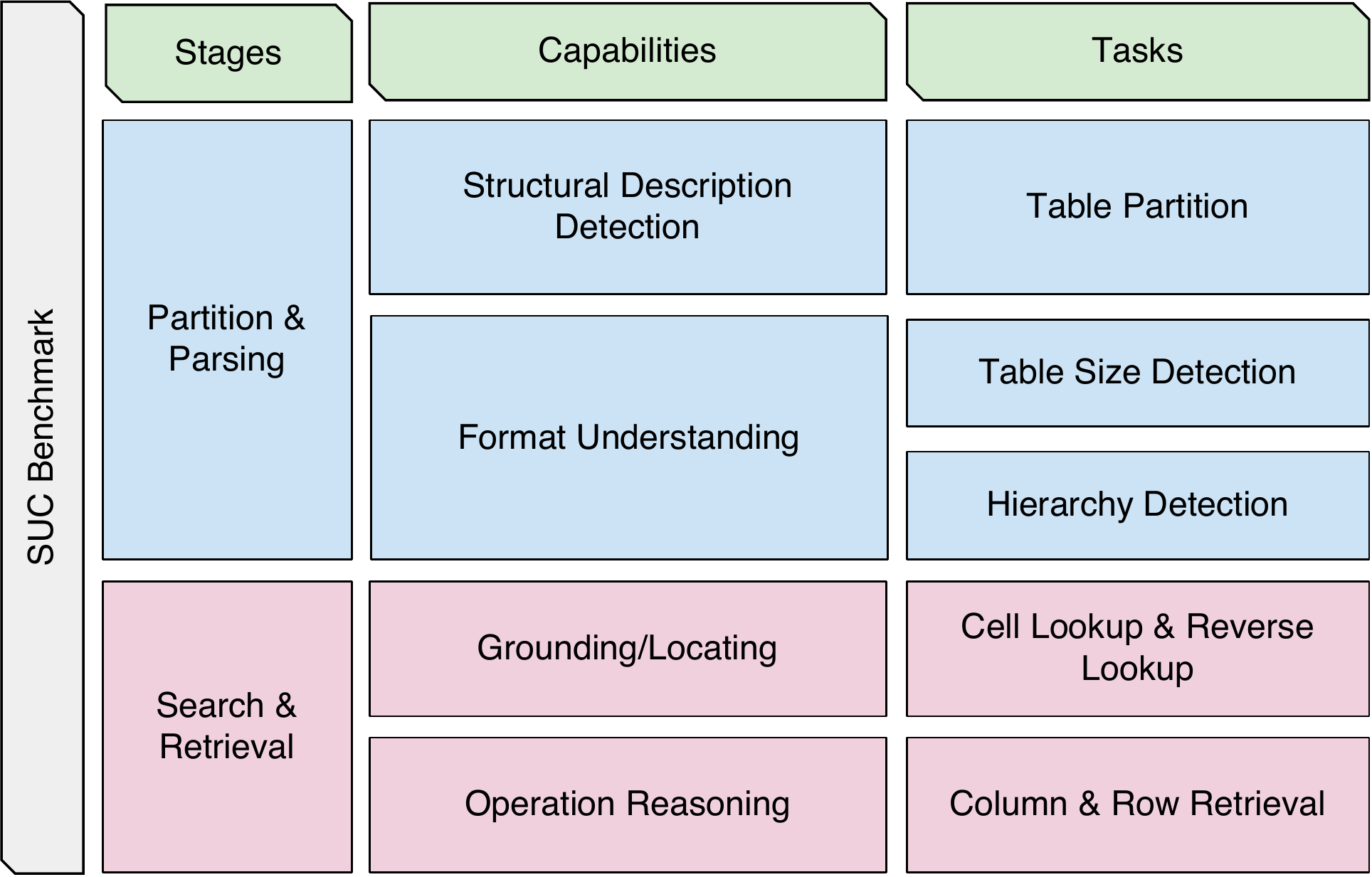}
    \caption{SUC Benchmark Overview}
    \label{fig:benchmark overview}
\end{figure}

We categorize the essential abilities to comprehend table structures from a human point of view into two distinct folds, as illustrated in Figure~\ref{fig:benchmark overview}.

\textit{1) Partition \& Parsing.} Tabular datasets are always paired with knowledge from other sources to provide more context and solve a specific downstream task. For example, HybridQA~\citep{dataset_hybridqa} employs passage information, TabFact~\citep{dataset_tabfact} and FEVEROUS~\citep{dataset_feverous} employs human annotation, and MultiModalQA~\citep{dataset_multimodalqa} employs image information.
However, the prerequisite for tackling these downstream tasks is the accurate partitioning of the data, which in turn requires the ability to distinguish tables from other supplementary information and an elementary understanding of the structural layout of tables. 

\begin{figure}[htbp]
    \centering
    \includegraphics[width=\linewidth]{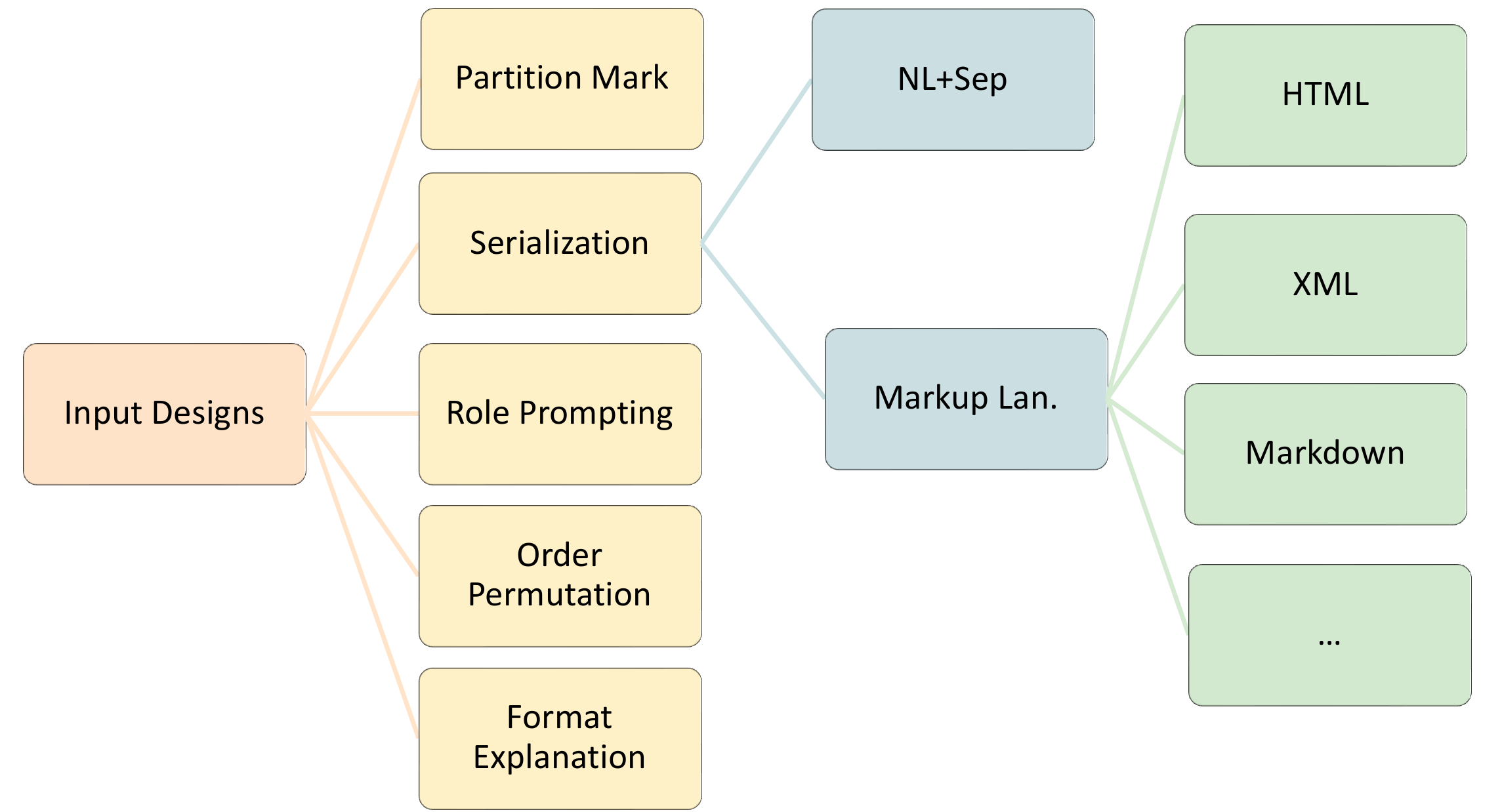}
    \caption{Input Designs for SUC Evaluation}
    \label{fig:input_designs}
\end{figure}

Additionally, various table storage formats, including CSV, JSON, XML, markdown, HTML~\citep{model_htlm} and XLSX, have different levels of information compression and present different challenges for LLMs in understanding the table content. For example, a table stored in CSV format is organized in rows with column values separated by commas, while a table stored in XML format is represented as a nested set of tags.
LLMs should first understand the format or layout of the table and then grasp its content. To our knowledge, no previous work has discussed the impact of these various storage formats. We aim to determine whether LLMs have the ability to correctly parse different formatting sources and identify which type of input design is most suitable for LLMs. It is also possible that LLMs already have the capability to handle all types of storage formats. The specific input designs can be found in Figure~\ref{fig:input_designs}.

\textit{2) Search \& Retrieval.} In addition to the capabilities mentioned above, the ability to accurately search and retrieve information from specific positions within structured data is crucial for LLMs. This capability is highly relevant to a wide range of downstream tasks, including but not limited to Table-QA and Column Type \& Relation Classification. It empowers LLMs to effectively identify and extract relevant information from structured data based on user queries or requests. For instance, consider a user asking, "For the Olympic events that took place after 2014, which event had an older flag bearer?" In order to answer this query, the LLM needs to first locate all the Olympic events that satisfy the time criterion, then compare the ages of the flag bearers associated with each event, and finally determine and return the event with the oldest flag bearer. The process of locating the relevant information within the structured data is achieved through careful analysis of the data's structure and the identification of the target cell or cells. By disentangling the search \& retrieval capabilities from the downstream tasks of LLMs, we gain valuable insights into the inner learning process of LLMs when it comes to tabular data.

\subsection{Task Design}
\label{sec:tasks} 

We have designed several specific tasks to assess the capabilities of LLMs in understanding tables (See concrete prompt design in Table~\ref{tab:input_design}). These tasks are designed in increasing difficulty.

\paragraph{Table Partition.} 
This task assesses the capability of LLMs to identify the structure of tables. The LLM is required to detect the boundaries of the tables within a given user input design. This input design may include various types of supplementary information, such as ``descriptions'', ``context'', ``statements'' and ``user queries''. Formally, given an input design, $D = {d_1, d_2, ...}$, where each part $d_{i}$ is a ``versatile'' sequence containing supplementary information such as description, context, statement, or user queries. For easy evaluation and comparison, we constrain LLMs to output a tuple of table boundary with the head token $b_h$ and the end token $b_e$ that includes the table content, as $B = (b_h, b_e)$.

\paragraph{Table Size Detection.} 
This task is essential to reveal the LLM's capability to correctly parse structural information. The size of a table is an important feature that is often overlooked. In fact, the table size feature represents direct constraints to how many rows and columns are encoded in a table. For instance, if a table only has three columns, the output should not consider answers outside this scope. Formally, given a table with $m$ rows and $n$ columns, a correct answer from a LLM should be $(m, n)$.

\paragraph{Merged Cell Detection.} 
This task assesses LLM's capability to parse structural information by detecting merged cells. Merged cells are special structures in table construction where two or more adjacent cells are combined to create a larger cell.
To test the robustness of LLMs, we consider merged cells as a feature in hierarchical spreadsheet tables. Formally, given a table with some merged cells, the LLM is required to detect the merged cell index $(r_i, c_j)$. Note that any index in the merged cell that matches the condition will be considered correct.

\paragraph{Cell Lookup \& Reverse Lookup.} 
This task reveals the capability to search and retrieve structural information. The LLM is required to accurately search and retrieve the cell value from a specific position. This task relies on the capabilities of information partitioning and parsing. In this task, if multiple cells with the same value are found, the LLM should retrieve their positions $(p_i, p_j), \cdots, (p_i^n, p_j^n)$. Conversely, given a specific cell position $(p_i, p_j)$, the LLM should retrieve the corresponding cell value $c_i$.

\paragraph{Column \& Row Retrieval.} 
This task assesses the LLM's capability to search and retrieve structural information by listing cell values. For column retrieval, the LLM is required to list the cell values $c_j$, $c_j^n$ of a specific column name $C_i$ from the given table. Similarly, for row retrieval, the LLM should list the cell values of a specific row index. In the evaluation process, we consider the prediction to be correct if the predicted value list matches the ground truth value list. We expect the performance of column/row retrieval tasks to be better than that of cell lookup and reverse lookup tasks, as using column/row indices to locate specific value lists is more common.

\begin{table}[htbp]
  \centering
  \caption{Input design of each task in our benchmark}
  \resizebox{\linewidth}{!}{
    \begin{tabular}{p{7.11em}p{23.445em}}
    \toprule
    \textbf{Task } & \textbf{Input} \\
    \midrule
    Table Partition & What is the first token (cell value instead of separator |) of the given table? What is the end token (cell value instead of separator |) of the given table? Answer questions one by one and use | to split the answer. \\
    \midrule
    Cell Lookup & What is the position of the cell value {cell\_value}? Use row index and column index to answer \\
    \midrule
    Reverse Lookup & What is the cell value of row index, column index ? Only output the cell value without other information \\
    \midrule
    Column Retrieval & What is the column name with the index {column\_idx} of the following table? Only give the column name without any explanation \\
    \midrule
    Row Retrieval & What are the cell values of the {row\_idx} row in following table? Only list the cell values one by one using | to split the answers \\
    \midrule
    Size Detection & How many rows in the table? How many columns in the table. Answer the questions one by one and use | to split the answer \\
    \midrule
    Merged Cell Detection & What is the column index of the cell which span is over 1. use | to split the answer (e.g., 3 | 4), the column index starts from 0. If there's no answer, return None \\
    \bottomrule
    \end{tabular}}
  \label{tab:input_design}%
\end{table}%

\subsection{Data Collection and Reformatting of SUC}
\label{sec:appendix-suc}

We collect structured data from various public datasets, \eg, TabFact~\citep{dataset_tabfact}, FEVEROUS~\citep{dataset_feverous}, SQA~\citep{dataset_sqa}, HybridQA~\citep{dataset_hybridqa} and ToTTo~\citep{dataset_totto}. All the tables are from Wikipedia. We only consider the structural portions of the original datasets, which are labeled with "table," "rows," or "headers," and exclude the other parts like "ID," "Answer," "Question," "FileName,". To identify a specific value within the structured data, we append each parsed sample with a unique question. Most of these questions are one sentence long, with a median length of 15 words. For example, ``How many rows (columns) are in the table?'' Each question is accompanied by a set of reference answers (``groundtruth'') sourced from the original datasets. For better evaluation, most of these questions are paired with some constraints such as "Answer the questions one by one and use "|" to split the answer.". We evaluate these questions using GPT-3.5 (Text-Davinci-003)\footnote{We perform the experiments through the public playground of OpenAI GPT-3.5 in \url{https://beta.openai.com/playground/}.} and manually eliminate any question that the model consistently answers correctly when multiple random samples are generated at a nonzero temperature\footnote{Temperature controls the randomness of the generation process. As the temperature approaches zero, the model becomes more deterministic and repetitive with very limited variation. Here, we set the temperature to 0.7 when creating the question and set the temperature to 0 when performing other experiments}. For the merged cell detection task, we only sample from ToTTo dataset since this is the only source paired with the merged cell. For each task setting, we randomly sample 1,500 tables for testing with a guaranteed table distribution.

\begin{figure*}[htbp]
  \centering
  \includegraphics[width=\linewidth]{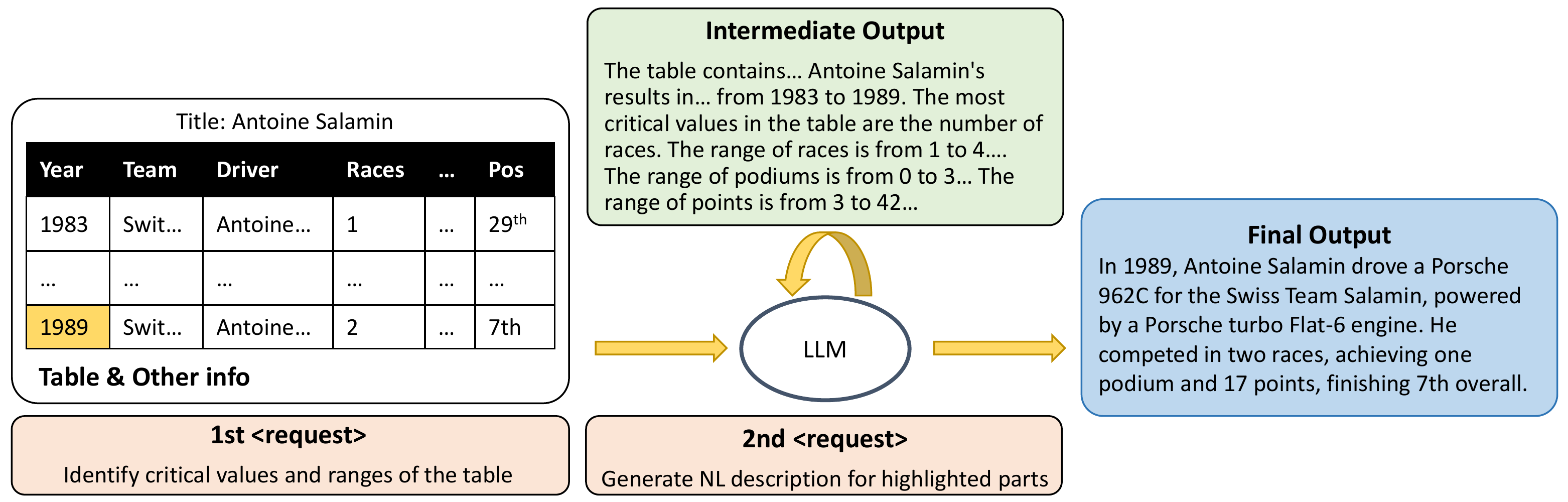}
  \caption{Illustration of self-augmented prompting. This process consists of two phases: 1) using self-augmented prompts to ask the LLM to generate additional knowledge (intermediate output) about the table; 2) incorporating the self-augmented response into the second prompt to request the final answer for a downstream task. As depicted in the figure, the LLM is able to identify important values in the table, which assists in generating a more accurate answer for the downstream task.}\label{pipeline}
\end{figure*}

\textit{One-shot Setting.} The SUC benchmark is designed as a one-shot in-context learning benchmark for tabular tasks. This means that the model can access one example from the SUC and may gain some context when generating the answers. Large Language Models (LLMs) have shown impressive capability in following few-shot prompts to accomplish unseen tasks without any fine-tuning~\cite{benchmark_truthfulqa}. This emergent capability is not captured by small language models. SUC leverages this property to better reveal the potential capabilities that LLMs may lack. We also conduct experiments using the zero-shot setting for comparison (See Table~\ref{tab:micro-ablation-input-designs}).

\subsection{Evaluation}
\label{sec:benchmark_evaluation}

We evaluate the benchmark using common input designs for table reasoning tasks and apply the methods to different LLMs for a deeper analysis. Specifically, we consider CSV, JSON, XML, markdown, HTML~\citep{model_htlm}, and XLSX as different format options. Each format represents a different level of information compression and poses different challenges for LLMs to understand the table content. we also consider using the most common way of concatenating a special token~\citep{model_tabert,model_unifiedSKG} as a separator, such as "|", as the baseline. The comparison numbers can be found in Table~\ref{tab:benchmark-micro}.
We also explore other input design options, such as grammar explanation, partition mark~\citep{model_uniSAr}, role prompting~\citep{model_tapex}, and format explanation, as augmentations for input designs. More details can be found in Figure~\ref{fig:input_designs} and Table~\ref{tab:micro-ablation-input-designs}.
In particular, we consider using the \textit{accuracy} for each task's evaluation.
To ensure better evaluation, we have added some constraints to the output format. For example, in the table partition task, we include the instruction "Answer questions one by one and use '|' to split the answer." Based on empirical observations, over 90\% of the answers follow these specific format instructions. For the remaining 10\% of samples, we apply a semantic-parsing strategy using regular expressions (Re)~\footnote{\url{https://docs.python.org/3/library/re.html}} to parse the answers.

\section{Structural Prompting}
\label{sec:prompting}

Our findings and insights over the SUC comparisons (See Sec~\refsec{sec:experiments}) have led us to the discovery that 1) LLMs have the basic structural understanding capabilities but are far from perfect, even on some trivial tasks, \eg, table size detection; 2) Correctly choosing the combination of input designs is a potential factor in boosting LLMs understanding over tabular data. In this section, we propose a simple and generic method, \textbf{self-augmented prompting}, to generate additional constraints using LLMs' self-knowledge. We find that when combined with carefully chosen input choices, these structural prompting methods lead to promising improvements on a variety of tabular downstream tasks (See Table~\ref{tab:downstream tasks}).

Recently, CoT~\citep{model_cot} has been discovered to empower LLMs to perform complex reasoning over text and lead to a long line of work~\citep{model_zero-shot-cot, model_verifiers, model_self_consistency, model_generateknowledge}. By providing the model with several exemplars of reasoning chains, LLMs can learn to follow the template to solve difficult unseen tasks. Inspired by these works, we propose a simple, generic, and effective method, \textit{self-augmented prompting}, to generate intermediate structural knowledge based on the internal retrieving of LLMs' self knowledge base. We design several ways to squeeze knowledge from LLM (see \reftab{tab:downstream tasks}). For example, we ask LLM to generate the format specification, which intends to clarify the input format pattern by LLM itself.
These choices diverge from previous approaches like CoT and Zero-shot-CoT~\citep{model_zero-shot-cot} by focusing on identifying effective methods for unlocking LLMs' capabilities to correctly comprehend structured information. Additionally, this method is \textit{model-agnostic}, that is, any standard structural data reasoning tasks can be used as the backbone, and can also be integrated with other prompting-based methods like self-consistency~\citep{model_self_consistency}.

\begin{table*}[htbp]
  \centering
    \caption{Micro results of the benchmark. Change order refers to place external text such as questions or statements before tables as described in~\citep{model_unifiedSKG}. "GPT-4" refers to the task performance of GPT-4 model. Given the resource-intensive nature of GPT-4 calls, we conduct inference test on a random subset of 300 samples from each task set. Each column in the results uses a graded color scale where deeper colors signify superior performance.}
  \label{tab:benchmark-micro}
  \resizebox{\textwidth}{!}{
    \begin{tabular}{lcccccccccccccc}
    \toprule
    \multicolumn{1}{c}{\multirow{2}[4]{*}{Format}} & \multicolumn{2}{c}{Table Partition} & \multicolumn{2}{c}{Cell Lookup} & \multicolumn{2}{c}{Reverse Lookup} & \multicolumn{2}{c}{Column Retrieval} & \multicolumn{2}{c}{Row Retrieval} & \multicolumn{2}{c}{Size Detection} & \multicolumn{2}{c}{Merged Cell Detection} \\
\cmidrule{2-15}          & Acc   & GPT-4 & Acc   & GPT-4 & Acc   & GPT-4 & Acc   & GPT-4 & Acc   & GPT-4 & Acc   & GPT-4 & Acc   & GPT-4 \\
    \midrule
    NL + Sep & \cellcolor[rgb]{ .988,  .988,  1}93.00\% & \cellcolor[rgb]{ .988,  .988,  1}96.78\% & \cellcolor[rgb]{ .988,  .988,  1}39.67\% & \cellcolor[rgb]{ .863,  .925,  .976}72.48\% & \cellcolor[rgb]{ .859,  .937,  .89}52.00\% & \cellcolor[rgb]{ .953,  .973,  .996}59.12\% & \cellcolor[rgb]{ .506,  .792,  .584}60.67\% & \cellcolor[rgb]{ .922,  .957,  .988}66.32\% & \cellcolor[rgb]{ .988,  .988,  1}31.00\% & \cellcolor[rgb]{ .988,  .988,  1}48.67\% & \cellcolor[rgb]{ .988,  .988,  1}42.00\% & \cellcolor[rgb]{ .988,  .988,  1}73.12\% & \cellcolor[rgb]{ .988,  .988,  1}71.33\% & \cellcolor[rgb]{ .988,  .988,  1}74.98\% \\
    Markdown & \cellcolor[rgb]{ .988,  .988,  1}92.33\% & \cellcolor[rgb]{ .796,  .89,  .961}\textbf{98.32\%} & \cellcolor[rgb]{ .576,  .824,  .643}43.33\% & \cellcolor[rgb]{ .906,  .949,  .984}71.93\% & \cellcolor[rgb]{ .988,  .988,  1}51.00\% & \cellcolor[rgb]{ .988,  .988,  1}57.32\% & \cellcolor[rgb]{ .988,  .988,  1}35.33\% & \cellcolor[rgb]{ .988,  .988,  1}60.12\% & \cellcolor[rgb]{ .388,  .745,  .482}\textbf{42.33\%} & \cellcolor[rgb]{ .855,  .922,  .976}49.98\% & \cellcolor[rgb]{ .988,  .988,  1}40.67\% & \cellcolor[rgb]{ .847,  .918,  .973}82.12\% & \cellcolor[rgb]{ .388,  .745,  .482}\textbf{78.00\%} & \cellcolor[rgb]{ .796,  .89,  .961}\textbf{82.64\%} \\
    JSON  & \cellcolor[rgb]{ .988,  .988,  1}94.00\% & \cellcolor[rgb]{ .988,  .988,  1}97.12\% & \cellcolor[rgb]{ .761,  .898,  .804}42.67\% & \cellcolor[rgb]{ .988,  .988,  1}68.32\% & \cellcolor[rgb]{ .494,  .788,  .573}54.33\% & \cellcolor[rgb]{ .988,  .988,  1}58.12\% & \cellcolor[rgb]{ .776,  .902,  .816}54.33\% & \cellcolor[rgb]{ .988,  .988,  1}64.32\% & \cellcolor[rgb]{ .988,  .988,  1}29.00\% & \cellcolor[rgb]{ .988,  .988,  1}48.32\% & \cellcolor[rgb]{ .988,  .988,  1}42.67\% & \cellcolor[rgb]{ .988,  .988,  1}76.43\% & \cellcolor[rgb]{ .988,  .988,  1}73.33\% & \cellcolor[rgb]{ .98,  .984,  1}78.98\% \\
     XML  & \cellcolor[rgb]{ .576,  .824,  .643}96.00\% & \cellcolor[rgb]{ .969,  .98,  .996}97.64\% & \cellcolor[rgb]{ .576,  .824,  .643}43.33\% & \cellcolor[rgb]{ .878,  .933,  .98}72.28\% & \cellcolor[rgb]{ .388,  .745,  .482}\textbf{55.00\%} & \cellcolor[rgb]{ .796,  .89,  .961}\textbf{60.32\%} & \cellcolor[rgb]{ .988,  .988,  1}41.33\% & \cellcolor[rgb]{ .843,  .914,  .973}68.28\% & \cellcolor[rgb]{ .51,  .796,  .588}41.00\% & \cellcolor[rgb]{ .796,  .89,  .961}\textbf{50.28\%} & \cellcolor[rgb]{ .988,  .988,  1}43.67\% & \cellcolor[rgb]{ .918,  .953,  .988}80.21\% & \cellcolor[rgb]{ .929,  .965,  .949}75.00\% & \cellcolor[rgb]{ .914,  .953,  .988}80.32\% \\
    HTML  & \cellcolor[rgb]{ .388,  .745,  .482}\textbf{96.67\%} & \cellcolor[rgb]{ .796,  .89,  .961}\textbf{98.32\%} & \cellcolor[rgb]{ .388,  .745,  .482}\textbf{44.00\%} & \cellcolor[rgb]{ .796,  .89,  .961}\textbf{73.34\%} & \cellcolor[rgb]{ .988,  .988,  1}47.33\% & \cellcolor[rgb]{ .91,  .949,  .984}59.45\% & \cellcolor[rgb]{ .388,  .745,  .482}\textbf{63.33\%} & \cellcolor[rgb]{ .796,  .89,  .961}\textbf{69.32\%} & \cellcolor[rgb]{ .42,  .761,  .51}42.00\% & \cellcolor[rgb]{ .816,  .902,  .965}50.19\% & \cellcolor[rgb]{ .388,  .745,  .482}\textbf{67.00\%} & \cellcolor[rgb]{ .796,  .89,  .961}\textbf{83.43\%} & \cellcolor[rgb]{ .631,  .843,  .69}76.67\% & \cellcolor[rgb]{ .867,  .925,  .976}81.28\% \\
    \bottomrule
    \end{tabular}}
\end{table*}

\begin{table*}[htbp]
  \centering
  \caption{Micro ablation results of the input designs over benchmark. All the $\Delta$ score denotes the performance gap compared to the first row, which includes all input design variants. Blue cells indicate an increase in performance (positive $\Delta$) while the red cells indicate an decrease in performance (negative $\Delta$). The intensity of the color corresponds to the magnitude of the change. The line marked with $^*$ indicates the highest average performance (65.43\%) across seven key tasks, achieved through the use of HTML markup with format explanations, role prompts, and by preserving the original sequence of inputs, \ie, placing external texts like questions and statements behind the tables. See detailed ablation results in Table~\ref{tab:benchmark}.}
  \label{tab:micro-ablation-input-designs}
  \resizebox{\textwidth}{!}{
    \begin{tabular}{lcccccccccccccccc}
    \toprule
    \multicolumn{1}{c}{\multirow{2}[3]{*}{Input Design}} & \multicolumn{2}{c}{Table Partition} & \multicolumn{2}{c}{Cell Lookup} & \multicolumn{2}{c}{Reverse Lookup} & \multicolumn{2}{c}{Column Retrieval} & \multicolumn{2}{c}{Row Retrieval} & \multicolumn{2}{c}{Size Detection} & \multicolumn{2}{c}{Merged Cell Detection} \\
\cmidrule{2-15}          & Acc   & $\Delta$     & Acc   & $\Delta$     & Acc   & $\Delta$     & Acc   & $\Delta$     & Acc   & $\Delta$     & Acc   & $\Delta$     & Acc   & $\Delta$ \\
\midrule
Markup Lang. HTML & 96.67\% & \cellcolor[rgb]{ .988,  .988,  1}0.00\% & 44.00\% & \cellcolor[rgb]{ .988,  .988,  1}0.00\% & 47.33\% & \cellcolor[rgb]{ .988,  .988,  1}0.00\% & 63.33\% & \cellcolor[rgb]{ .988,  .988,  1}0.00\% & 42.00\% & \cellcolor[rgb]{ .988,  .988,  1}0.00\% & 67.00\% & \cellcolor[rgb]{ .988,  .988,  1}0.00\% & 76.67\% & \cellcolor[rgb]{ .988,  .988,  1}0.00\% \\
    w/o format explanation & 92.00\% & \cellcolor[rgb]{ .98,  .808,  .816}-4.67\% & 52.00\% & \cellcolor[rgb]{ .651,  .753,  .882}8.00\% & 52.33\% & \cellcolor[rgb]{ .78,  .843,  .929}5.00\% & 64.33\% & \cellcolor[rgb]{ .949,  .961,  .988}1.00\% & 36.00\% & \cellcolor[rgb]{ .98,  .757,  .765}-6.00\% & 78.00\% & \cellcolor[rgb]{ .525,  .663,  .839}11.00\% & 77.67\% & \cellcolor[rgb]{ .949,  .961,  .988}1.00\% \\
   \textbf{w/o partition mark$^*$} & 98.00\% & \cellcolor[rgb]{ .933,  .949,  .98}1.33\% & 59.00\% & \cellcolor[rgb]{ .357,  .545,  .78}15.00\% & 53.00\% & \cellcolor[rgb]{ .749,  .82,  .918}5.67\% & 66.00\% & \cellcolor[rgb]{ .878,  .91,  .961}2.67\% & 39.67\% & \cellcolor[rgb]{ .984,  .898,  .906}-2.33\% & 72.00\% & \cellcolor[rgb]{ .78,  .843,  .929}5.00\% & 70.33\% & \cellcolor[rgb]{ .98,  .741,  .753}-6.33\% \\
    w/o role prompting & 95.00\% & \cellcolor[rgb]{ .984,  .922,  .933}-1.67\% & 40.67\% & \cellcolor[rgb]{ .984,  .859,  .871}-3.33\% & 44.67\% & \cellcolor[rgb]{ .984,  .882,  .894}-2.67\% & 59.00\% & \cellcolor[rgb]{ .98,  .82,  .831}-4.33\% & 39.33\% & \cellcolor[rgb]{ .984,  .882,  .894}-2.67\% & 69.00\% & \cellcolor[rgb]{ .906,  .929,  .973}2.00\% & 76.00\% & \cellcolor[rgb]{ .984,  .961,  .973}-0.67\% \\
    w/o change order & 96.67\% & \cellcolor[rgb]{ .988,  .988,  1}0.00\% & 52.33\% & \cellcolor[rgb]{ .639,  .741,  .878}8.33\% & 40.67\% & \cellcolor[rgb]{ .98,  .729,  .741}-6.67\% & 55.67\% & \cellcolor[rgb]{ .976,  .69,  .702}-7.67\% & 31.67\% & \cellcolor[rgb]{ .976,  .588,  .6}-10.33\% & 52.67\% & \cellcolor[rgb]{ .973,  .435,  .443}-14.33\% & 65.67\% & \cellcolor[rgb]{ .976,  .565,  .573}-11.00\% \\
\midrule
w/o 1-shot & 63.00\% & \cellcolor[rgb]{ .973,  .412,  .42}-33.67\% & 9.33\% & \cellcolor[rgb]{ .973,  .412,  .42}-34.67\% & 17.33\% & \cellcolor[rgb]{ .973,  .412,  .42}-30.00\% & 50.00\% & \cellcolor[rgb]{ .973,  .475,  .482}-13.33\% & 30.00\% & \cellcolor[rgb]{ .973,  .525,  .533}-12.00\% & 16.67\% & \cellcolor[rgb]{ .973,  .412,  .42}-50.33\% & 38.00\% & \cellcolor[rgb]{ .973,  .412,  .42}-38.67\% 
\\
    \bottomrule
    \end{tabular}
    }
\end{table*}

Self-augmented prompting is a straightforward idea that invokes a LLM twice to enhance its ability to understand structured data, as illustrated in Figure~\ref{pipeline}. 
This method leverages the LLM's capacity to extract and process complex data by breaking down the task into two distinct phases.
In the first stage, the model is prompted to focus on identifying crucial values and patterns in the data. Instead of directly asking the LLM to solve the primary task, it is initially tasked with extracting significant data points or determining ranges that are relevant to the downstream process. This is crucial because these extracted elements are key to building a better understanding of the context and the specific details needed for more accurate results.
Once the relevant data has been identified and outlined by the LLM, the second stage involves using this refined data as part of a new prompt that now focuses on the original task. By incorporating the results from the first prompt into the second, the process effectively guides the LLM to consider these critical details while generating the final answer, enhancing both the relevance and accuracy of the response.

The merit of this approach lies in its capacity to transform a general query into a structured two-step inquiry, optimizing the model's analytical prowess before applying it to the actual problem-solving task. This not only streamlines the reasoning process but also significantly improves the outcome by structuring the interaction with the LLM around the specific insights needed to address the query comprehensively.
Refer to Table~\ref{tab:downstream tasks} for a comparison of self-augmented prompting experimental results.

\begin{table*}[htbp]
  \centering
  \caption{Comparison of self-augmented and 1-shot prompting across various design variants on downstream tasks. Refer to the table~\ref{tab:instruction} for details on different self-augmented prompting options.}
  \label{tab:downstream tasks}
  \resizebox{\textwidth}{!}{
    \begin{tabular}{clcccccccc}
    \toprule
    \multirow{2}[4]{*}{Type} & \multicolumn{1}{c}{\multirow{2}[4]{*}{Choice}} & TabFact & HybridQA & SQA   & Feverous & \multicolumn{4}{c}{ToTTo} \\
\cmidrule{3-10}          &       & Acc   & Acc   & Acc   & Acc   & BLEU-1 & BLEU-2 & BLEU-3 & BLEU-4 \\
    \midrule
    1-shot & 1-shot & \cellcolor[rgb]{ .847,  .882,  .953}72.04\% & \cellcolor[rgb]{ .886,  .91,  .969}46.07\% & \cellcolor[rgb]{ .8,  .847,  .941}73.81\% & \cellcolor[rgb]{ .843,  .878,  .953}75.56\% & \cellcolor[rgb]{ .91,  .929,  .976}72.43\% & \cellcolor[rgb]{ .878,  .906,  .965}44.36\% & \cellcolor[rgb]{ .898,  .922,  .973}27.01\% & \cellcolor[rgb]{ .922,  .937,  .98}17.24\% \\
    1-shot & w/o table size & \cellcolor[rgb]{ .933,  .945,  .984}71.33\% & \cellcolor[rgb]{ .969,  .976,  .996}45.52\% & \cellcolor[rgb]{ .882,  .91,  .965}72.91\% & \cellcolor[rgb]{ .988,  .988,  1}74.66\% & \cellcolor[rgb]{ .918,  .933,  .976}72.30\% & \cellcolor[rgb]{ .89,  .914,  .969}44.23\% & \cellcolor[rgb]{ .886,  .914,  .969}27.14\% & \cellcolor[rgb]{ .922,  .937,  .98}17.25\% \\
    1-shot & w/o partition mark & \cellcolor[rgb]{ .941,  .953,  .988}71.25\% & \cellcolor[rgb]{ .976,  .98,  .996}45.48\% & \cellcolor[rgb]{ .867,  .894,  .961}73.09\% & \cellcolor[rgb]{ .914,  .933,  .976}75.11\% & \cellcolor[rgb]{ .965,  .969,  .992}71.18\% & \cellcolor[rgb]{ .984,  .984,  1}43.17\% & \cellcolor[rgb]{ .957,  .965,  .992}26.36\% & \cellcolor[rgb]{ .976,  .98,  .996}16.34\% \\
    1-shot & w/o format explanation & \cellcolor[rgb]{ .988,  .988,  1}70.87\% & \cellcolor[rgb]{ .988,  .988,  1}45.39\% & \cellcolor[rgb]{ .988,  .988,  1}71.69\% & \cellcolor[rgb]{ .773,  .827,  .929}75.97\% & \cellcolor[rgb]{ .988,  .988,  1}70.54\% & \cellcolor[rgb]{ .945,  .957,  .988}43.59\% & \cellcolor[rgb]{ .945,  .957,  .988}26.52\% & \cellcolor[rgb]{ .953,  .961,  .988}16.74\% \\
    1-shot & w/o role prompting & \cellcolor[rgb]{ .929,  .945,  .984}71.35\% & \cellcolor[rgb]{ .89,  .914,  .969}46.05\% & \cellcolor[rgb]{ .839,  .875,  .953}73.39\% & \cellcolor[rgb]{ .847,  .882,  .957}75.52\% & \cellcolor[rgb]{ .988,  .988,  1}70.61\% & \cellcolor[rgb]{ .988,  .988,  1}43.10\% & \cellcolor[rgb]{ .988,  .988,  1}26.02\% & \cellcolor[rgb]{ .988,  .988,  1}16.15\% \\
    \midrule
    SA    & self format explanation & \cellcolor[rgb]{ .824,  .863,  .945}72.23\% & \cellcolor[rgb]{ .878,  .906,  .965}46.12\% & \cellcolor[rgb]{ .792,  .839,  .937}73.91\% & \cellcolor[rgb]{ .745,  .804,  .922}76.15\% & \cellcolor[rgb]{ .839,  .875,  .953}74.18\% & \cellcolor[rgb]{ .8,  .847,  .941}45.25\% & \cellcolor[rgb]{ .871,  .898,  .961}27.32\% & \cellcolor[rgb]{ .851,  .886,  .957}18.34\% \\
    SA    & self critical values and ranges identification & \cellcolor[rgb]{ .557,  .663,  .859}74.35\% & \cellcolor[rgb]{ .557,  .663,  .859}48.20\% & \cellcolor[rgb]{ .557,  .663,  .859}76.53\% & \cellcolor[rgb]{ .718,  .784,  .914}76.32\% & \cellcolor[rgb]{ .557,  .663,  .859}80.83\% & \cellcolor[rgb]{ .557,  .663,  .859}47.96\% & \cellcolor[rgb]{ .557,  .663,  .859}30.68\% & \cellcolor[rgb]{ .557,  .663,  .859}22.92\% \\
    SA    & self structural information description & \cellcolor[rgb]{ .675,  .753,  .898}73.42\% & \cellcolor[rgb]{ .749,  .808,  .922}46.97\% & \cellcolor[rgb]{ .608,  .702,  .878}75.97\% & \cellcolor[rgb]{ .557,  .663,  .859}77.28\% & \cellcolor[rgb]{ .639,  .725,  .886}78.93\% & \cellcolor[rgb]{ .651,  .733,  .89}46.91\% & \cellcolor[rgb]{ .722,  .784,  .914}28.94\% & \cellcolor[rgb]{ .788,  .839,  .937}19.32\% \\
    \bottomrule
    \end{tabular}}
\end{table*}
\begin{table}[htbp]
  \centering
  \caption{Main results of the downstream tasks ablation study}
  \label{tab:downstreams_ablation_micro}
  \resizebox{\columnwidth}{!}{
    \begin{tabular}{lccccc}
    \toprule
    \multicolumn{1}{c}{\multirow{2}[4]{*}{Format}} & TabFact & HybridQA & SQA   & Feverous & ToTTo \\
\cmidrule{2-6}          & Acc   & Acc   & Acc   & Acc   & BLEU-4 \\
    \midrule
    NL + Sep & \cellcolor[rgb]{ .78,  .906,  .82}70.26\% & \cellcolor[rgb]{ .945,  .973,  .965}45.02\% & \cellcolor[rgb]{ .62,  .839,  .682}70.41\% & \cellcolor[rgb]{ .408,  .753,  .498}75.15\% & \cellcolor[rgb]{ .388,  .745,  .482}\textbf{12.70\%} \\
    Markdown & \cellcolor[rgb]{ .988,  .988,  1}68.40\% & \cellcolor[rgb]{ .733,  .886,  .78}45.88\% & \cellcolor[rgb]{ .988,  .988,  1}66.59\% & \cellcolor[rgb]{ .988,  .988,  1}71.88\% & \cellcolor[rgb]{ .988,  .988,  1}8.57\% \\
    JSON  & \cellcolor[rgb]{ .988,  .988,  1}68.04\% & \cellcolor[rgb]{ .988,  .988,  1}42.40\% & \cellcolor[rgb]{ .624,  .843,  .686}70.39\% & \cellcolor[rgb]{ .882,  .945,  .91}73.84\% & \cellcolor[rgb]{ .988,  .988,  1}8.82\% \\
     XML  & \cellcolor[rgb]{ .875,  .945,  .902}70.00\% & \cellcolor[rgb]{ .412,  .757,  .502}47.20\% & \cellcolor[rgb]{ .533,  .804,  .608}70.74\% & \cellcolor[rgb]{ .988,  .988,  1}73.14\% & \cellcolor[rgb]{ .988,  .988,  1}8.82\% \\
    HTML  & \cellcolor[rgb]{ .388,  .745,  .482}\textbf{71.33\%} & \cellcolor[rgb]{ .388,  .745,  .482}\textbf{47.29\%} & \cellcolor[rgb]{ .388,  .745,  .482}\textbf{71.31\%} & \cellcolor[rgb]{ .388,  .745,  .482}\textbf{75.20\%} & \cellcolor[rgb]{ .506,  .792,  .584}12.30\% \\
    \midrule
    GPT-4 w/ HTML & 78.40\% & 56.68\% & 75.35\% & 83.21\% & 20.12\% \\
    \bottomrule
    \end{tabular}}
\end{table}

Based on the empirical observations, it is evident that structural information plays a crucial role in comprehending a table. Researchers such as \citet{model_unifiedSKG, model_tabert, model_htlm} have made progress by incorporating prompts with special tokens to encode different structural information. Building on their work and the findings from the SUC benchmark results, we explore the concept of manual prompt engineering as an additional technique for self-augmented prompting.
Specifically, we consider extracting structural information from the raw input and incorporating it into the input itself. This can involve using cell addresses and clearly indicating the number of rows and columns in the table. Such augmentation aims to provide additional knowledge and constraints, thereby improving the LLM's ability to reason in tabular downstream tasks.
We have observed that the LLM performs poorly in the task of table size detection (refer to Section~\ref{sec:benchmark}), which motivates us to include structural features in the input. For example, we append information about the table size and merged cell positions to create a more structure-aware in-context learning environment for downstream tasks. Our ablation study in Sec~\refsec{benchmark_high_lights} shows that appending table size and merged cell position leads to an improvement in the LLM's performance on downstream tasks.

\begin{table*}[htbp]
  \centering
  \caption{Instruction for self-augmented prompting in Section~\ref{sec:prompting}.}
  \resizebox{0.85\linewidth}{!}{
    \begin{tabular}{p{16.11em}p{35.445em}}
    \toprule
    \textbf{Method } & \textbf{Instruction} \\
    \midrule
    Format explanation & Generate short format specification and description of the table  within five sentences.\\
    \midrule
    Critical values and ranges identification & Identify critical values and ranges of the table related within five sentences. \\
    \midrule
    Structural information description & Describe structural information, patterns and statistics of the table within five sentences.\\
    \bottomrule
    \end{tabular}
    }
  \label{tab:instruction}%
\end{table*}%

\section{Experiments}
\label{sec:experiments}

\subsection{Experiment Settings}

\textbf{Models.}
In this study, we evaluate the performance on GPT-3.5~\citep{model_instruct_gpt3} and GPT-4~\citep{openaiGPT4TechnicalReport2023}. Unless otherwise specified, we utilize text-davinci-003 in all experiments. Specifically, we set the hyper-parameter temperature to 0, top\_p to 1, with n set to 1 when performing the experiments;
\textbf{Downstream Tasks and Datasets.}
In addition to evaluate LLMs' capabilities toward understanding structured data through our benchmark. We also conduct experiments on five typical tabular downstream tasks. The datasets are shown as follows, and the evaluation number can be found in Table~\ref{tab:downstream tasks}.

Specifically, we use \textbf{(1) SQA} which is composed of 6,066 question sequences (2.9 questions per sequence on average), constructed by decomposing a subset of highly compositional WTQ questions; \textbf{(2) HybridQA} which requires reasoning on heterogeneous information rather than homogeneous information alone, which involves 62,682 questions. Each question is aligned with a Wikipedia table and multiple free-form corpora linked with the entities in the table. The questions are designed to aggregate both tabular information and text information, \ie, lack of either form would render the question unanswerable; \textbf{(3) ToTTo} which is a high-quality English table-to-text dataset with more than 100,000 examples in which a table from Wikipedia with highlighted cells is paired with a sentence that describes the highlighted cells. The task is like given a Wikipedia table with row names, column names and table cells, with a subset of cells highlighted, generate a natural language description for the highlighted part of the table; \textbf{(4) FEVEROUS} which is a fact verification dataset consisting of 87,026 verified claims. Each claim is annotated with evidence in the form of sentences and/or cells from tables in Wikipedia, as well as a label indicating whether this evidence supports, refutes, or does not provide enough information to reach a verdict;
\textbf{(5) TabFact} which is a fact verification dataset in which the tables were extracted from Wikipedia and sentences were written by crowd workers.

\subsection{Results}

\subsubsection{Benchmark Highlights}
\label{benchmark_high_lights}

Comprehensive evaluations of different structural understanding tasks with various input designs over SUC are presented in Table~\ref{tab:micro-ablation-input-designs}. 
The results show that the system's overall accuracy gets highest when using the HTML markup language with format explanations and role prompts, and without order change, achieving a 65.43\% overall accuracy on seven tasks. It indicates that the LLM has significant potential for understanding the structural information of tables in this specific format. However, it is also evident that the LLM's performance is negatively impacted when certain features are removed, especially when the prompt example is removed.
We give some highlights associated with the benchmark results as follows:

\paragraph{NL+Sep vs. Markup Lan.} 
We compare the use of natural language with specific separators (NL+Sep) and markup languages such as HTML, XML, and JSON. 
Even ``NL+Sep'' is commonly used in tabular downstream tasks~\citep{model_tabert, model_tapas, model_tapex, model_table_reasoner}, however, our results show that using markup languages, specifically HTML, outperforms ``NL+Sep'' with a 6.76\% improvement.
We assume that the training process of the LLMs involves code tuning and that the training dataset contains a significant amount of web data. As a result, the LLM is more familiar with HTML and XML formats when interpreting tables. (For more information about the GPT-3.5 training, see~\citep{model_instruct_gpt3}).

\paragraph{1-shot vs. 0-shot.} A notable finding is that the system's performance drops significantly when it is in a zero-shot setting, with an overall accuracy decrease of 30.38\% on all tasks using HTML format. This indicates that learning structural information is highly dependent on in-context learning. This is particularly significant for tasks such as size detection and merged cell detection, which are closely related to the ability to parse structural information.

\paragraph{External information should appear ahead of tables.} In order to understand the impact of the order on the input design, we observed that when we manually placed external information such as questions and statements behind the table, there was an overall 6.81\% decrease in performance across all tasks. One possible explanation for this is that placing external information ahead of tables could assist LLM in better generalization and gaining more context regarding the structural information of tables.

\paragraph{Partition mark and format explanation may undermine Search and Retrieval capability}
Partition mark~\citep{model_uniSAr} is commonly used in input designs. Inspired by the partition mark, we propose another similar choice called "format explanation". It provides an additional explanation of the adopted format. For example, in the case of HTML format, we explain that "Each table cell is defined by a <td> and a </td> tag; Each table row starts with a <tr> and ends with a </tr> tag; th stands for table header." However, when it comes to the task of Cell Lookup, adding partition marks and format explanations actually results in a decrease in performance across all input designs. This suggests that such additional structural information may bias the searching and retrieval process of LLM over the tabular structure. However, adding partition marks or format explanations does show some benefits for specific tasks such as merged cell detection. To provide a clearer understanding of the impact of adding additional explanations or special tokens, we conducted experiments as shown in Table~\ref{tab:downstream tasks}. The results reveal that while they may undermine the search and retrieval capability of LLMs, they still improve overall performance in downstream tasks.

The SUC benchmark provides a comprehensive comparison using multiple input designs to evaluate the structural understanding capabilities of tables. Based on the findings, guidelines are proposed to address the questions mentioned in Sec~\refsec{sec:benchmark}:
\begin{itemize}
    \item 1) LLMs have basic structural understanding capabilities, but are far from perfect, even for some trivial tasks, \eg, table size detection; 
    \item 2) Choosing the right combination of input designs is a potential factor in improving the performance of LLMs when working with structured data.
\end{itemize}

\subsubsection{Downstream Tasks}

Table~\ref{tab:downstream tasks} gives the comparison using self-augmented prompting with the trick of structural feature detection. We find one notable insight that the models perform better using self-augmented prompting than 1-shot, as seen in the "SA" rows where the models are given self-generated information: format explanation, key range and values identification, and structural information description. These rows have higher accuracy and BLEU scores than the "1-shot" rows. Especially for generating key range and values identification gives an overall 3.26\% performance improvement than 1-shot setting on all the tasks. Furthermore, manually prompting structural features like ``table size'' from the input has a slight impact on the model's performance on downstream tasks. The drop of table size option gives an overall 0.4\% decrease, as seen by the small difference between the ``1-shot'' row and the ``1-shot w/o table size'' row.

\begin{table}[htbp]
  \centering
  \caption{Format explanation example from FEVEROUS}
  \label{tab:format-example}
  \resizebox{\linewidth}{!}{
    \begin{tabular}{p{7.39em}p{17.78em}}
    \toprule
    \textbf{Manual Prompt} & \textbf{Self-augmented Prompt Generation} \\
    \midrule
    Each table cell is defined by a <td> and a </td> tag.\newline{} Each table row starts with a <tr> and ends with a </tr> tag.\newline{} th stands for table header. & '15 rows and 4 columns, with the first column being an index, the second column being empty, the third column being the name of the Marshal, the fourth column being the date of promotion, and the fifth column being the defence branch.\textbackslash{}nThe table is defined by HTML tags, with each table cell being defined by a <td> and a </td> tag, and each table row starting with a <tr> and ending with a </tr> tag.\textbackslash{}nThe table header is denoted by the th tag.' \\
    \bottomrule
    \end{tabular}}
\end{table}

Based on the comparison between the "1-shot w/o format explanation" row and the "SA self format explanation" row, we observe that manually labeling the format explanation may have a negative impact on downstream tasks like FEVEROUS. This is because the table structure in FEVEROUS is more irregular, with numerous segments and subtables. These structural complexities pose significant challenges for GPT-3.5. Additionally, manually-crafted knowledge is more general and cannot cover detailed information of this nature. On the other hand, self-augmented prompting can learn patterns independently and generate more comprehensive and helpful cues to address the questions. We provide an example of a format explanation from FEVEROUS using two prompt designs in Table~\ref{tab:format-example}.

\section{Related Work}
\label{sec:related_work}

\textbf{In-context Learning with LLMs.}
Large language models, such as GPT-3~\citep{model_gpt3}, Instruct-GPT, and Codex~\citep{model_codex}, have demonstrated their capability as few-shot reasoners in natural language-related tasks. The effectiveness of this capability is influenced by factors such as the model size, the amount of data used, and the available computing power. Recent studies~\cite{model_flan_t5, model_verifiers, model_scaling} have proposed various methods for training these large language models (LLMs). These models have exhibited an impressive ability to perform tasks that they haven't been specifically fine-tuned for, which is an emergent capability not observed in smaller language models.

\textbf{Intermediate of Prompt Engineering.}
Recently, several intermediate prompt engineering methods have been proposed following "CoT"~\citep{model_cot}. CoT provides a few examples with explanations of the reasoning process. This step-by-step reasoning approach helps LLMs generate more accurate results. However, according to \citep{model_cot}, "CoT only yields performance gains when used with models of nearly 100B parameters." Smaller models tend to produce illogical chains of thought, resulting in lower accuracy compared to standard prompting. Typically, the performance boost from CoT prompting is proportional to the model's size.
Zero-shot chain of Thought (Zero-shot-CoT) prompting is a follow-up to CoT that introduces an incredibly simple zero-shot prompt. By appending the words "Let's think step by step." to the end of a question, LLMs can generate a chain of thoughts that answers the question. Extracting answers from this chain of thought leads to more accurate results. Another follow-up to CoT is Self-consistency~\citep{model_self_consistency}, which generates multiple chains of thoughts and selects the majority answer based on a voting strategy as the final answer. Self-consistency has shown improvements in arithmetic, commonsense, and symbolic reasoning tasks. Even when regular CoT is found to be ineffective, self-consistency can still enhance results.
In addition, \citet{model_generateknowledge} propose the generated knowledge approach, which prompts the LLM to generate potentially useful information about the question before generating a response.

\section{Conclusion}
\label{sec:conclusion}

In this paper, we propose a benchmark to compare various input designs in order to study the structural understanding capabilities of LLMs on tables. Surprisingly, we obtain some insights of the input designs and the comparison reveal that LLMs have the basic capabilities towards understanding structural information of tables. We also give some guidance on how to apply our benchmark insights on downstream tasks and propose a simple, generic but effective method, \ie, self-augmented prompting, by generating additional knowledge with LLMs self-knowledge. We believe this study will be beneficial for table-based, even structured data based research, or serve as a auxiliary tool to help better understand the table(s) from structural perspectives.

\section*{Ethical Considerations}
\label{sec:limitations}

Structured data often includes metadata, which provides additional information about the data and helps to provide context (\eg, column names, data types, \etc). Interpreting and utilizing metadata is a challenge when the meaning and significance of the structured data may not be immediately apparent and must be inferred from the metadata and other contextual clues. This capability is highly dependent on downstream tasks, such as column type prediction~\citep{dataset_gittables} and dimension/measure classification~\citep{model_metadata}. We believe that understanding this challenge is an important area of research. However, due to space limitations, we will leave this section for further exploration.
Furthermore, our method is primarily designed for languages with limited morphology, such as English. The scalability of our approach to longer texts is a topic that we will explore in more detail.


\newpage
\bibliographystyle{utils/acm-reference-format}
\bibliography{references}

\appendix
\section{Appendix}
\label{sec:appendix}

\begin{table*}[htbp]
  \centering
  \caption{Full results of the downstream tasks.}
  \label{tab:downstreams_ablation}
  \resizebox{\textwidth}{!}{
    \begin{tabular}{lcccccccccccccccc}
    \toprule
    \multicolumn{1}{c}{\multirow{2}[4]{*}{Input Design}} & \multicolumn{2}{c}{TabFact} & \multicolumn{2}{c}{HybridQA} & \multicolumn{2}{c}{SQA} & \multicolumn{2}{c}{Feverous} & \multicolumn{8}{c}{ToTTo} \\
\cmidrule{2-17}          & Acc   & $\Delta$     & Acc   & $\Delta$     & Acc   & $\Delta$     & Acc   & $\Delta$     & BLEU-1 & $\Delta$     & BLEU-2 & $\Delta$     & BLEU-3 & $\Delta$     & BLEU-4 & $\Delta$ \\
    \midrule
    \textbf{NL + Sep} & 70.26\% & \cellcolor[rgb]{ .988,  .988,  1}0.00\% & 45.02\% & \cellcolor[rgb]{ .988,  .988,  1}0.00\% & 70.41\% & \cellcolor[rgb]{ .988,  .988,  1}0.00\% & \textbf{75.15\%} & \cellcolor[rgb]{ .988,  .988,  1}0.00\% & 67.64\% & \cellcolor[rgb]{ .988,  .988,  1}0.00\% & 38.25\% & \cellcolor[rgb]{ .988,  .988,  1}0.00\% & 21.28\% & \cellcolor[rgb]{ .988,  .988,  1}0.00\% & 12.70\% & \cellcolor[rgb]{ .988,  .988,  1}0.00\% \\
        w/o format explanation & 69.61\% & \cellcolor[rgb]{ .984,  .961,  .973}-0.65\% & 44.28\% & \cellcolor[rgb]{ .984,  .957,  .969}-0.74\% & 70.92\% & \cellcolor[rgb]{ .969,  .976,  .996}0.51\% & 71.03\% & \cellcolor[rgb]{ .98,  .827,  .839}-4.12\% & 72.49\% & \cellcolor[rgb]{ .784,  .847,  .929}4.84\% & \textbf{42.50\%} & \cellcolor[rgb]{ .812,  .863,  .937}4.25\% & \textbf{24.32\%} & \cellcolor[rgb]{ .863,  .898,  .957}3.04\% & \textbf{13.71\%} & \cellcolor[rgb]{ .949,  .961,  .988}1.01\% \\
        w/o partition mark & 69.12\% & \cellcolor[rgb]{ .984,  .941,  .953}-1.14\% & 43.68\% & \cellcolor[rgb]{ .984,  .933,  .945}-1.35\% & \textbf{71.79\%} & \cellcolor[rgb]{ .933,  .949,  .98}1.38\% & \textbf{74.37\%} & \cellcolor[rgb]{ .984,  .957,  .969}-0.78\% & 73.29\% & \cellcolor[rgb]{ .749,  .824,  .918}5.65\% & \textbf{40.99\%} & \cellcolor[rgb]{ .875,  .91,  .961}2.73\% & \textbf{22.58\%} & \cellcolor[rgb]{ .937,  .953,  .984}1.29\% & 13.28\% & \cellcolor[rgb]{ .965,  .973,  .992}0.58\% \\
        w/o role prompting & 68.89\% & \cellcolor[rgb]{ .984,  .933,  .945}-1.37\% & 42.88\% & \cellcolor[rgb]{ .984,  .902,  .914}-2.14\% & 70.74\% & \cellcolor[rgb]{ .976,  .98,  .996}0.33\% & 73.71\% & \cellcolor[rgb]{ .984,  .929,  .941}-1.45\% & 48.58\% & \cellcolor[rgb]{ .973,  .412,  .42}-19.06\% & 28.49\% & \cellcolor[rgb]{ .976,  .612,  .62}-9.76\% & 16.35\% & \cellcolor[rgb]{ .98,  .796,  .808}-4.93\% & 9.83\% & \cellcolor[rgb]{ .984,  .875,  .886}-2.87\% \\
    \midrule
    \textbf{Markup Lan. Markdown} & 68.40\% & \cellcolor[rgb]{ .988,  .988,  1}0.00\% & \textbf{45.88\%} & \cellcolor[rgb]{ .988,  .988,  1}0.00\% & 66.59\% & \cellcolor[rgb]{ .988,  .988,  1}0.00\% & 71.88\% & \cellcolor[rgb]{ .988,  .988,  1}0.00\% & 63.96\% & \cellcolor[rgb]{ .988,  .988,  1}0.00\% & 33.44\% & \cellcolor[rgb]{ .988,  .988,  1}0.00\% & 15.82\% & \cellcolor[rgb]{ .988,  .988,  1}0.00\% & 8.57\% & \cellcolor[rgb]{ .988,  .988,  1}0.00\% \\
        w/o format explanation & 67.90\% & \cellcolor[rgb]{ .984,  .969,  .98}-0.51\% & 43.09\% & \cellcolor[rgb]{ .984,  .878,  .89}-2.80\% & 66.96\% & \cellcolor[rgb]{ .976,  .98,  .996}0.36\% & 65.97\% & \cellcolor[rgb]{ .98,  .761,  .769}-5.91\% & 70.41\% & \cellcolor[rgb]{ .718,  .8,  .906}6.45\% & 38.79\% & \cellcolor[rgb]{ .765,  .831,  .922}5.35\% & 18.71\% & \cellcolor[rgb]{ .867,  .906,  .961}2.89\% & \textbf{18.71\%} & \cellcolor[rgb]{ .561,  .686,  .851}10.14\% \\
        w/o partition mark & 67.53\% & \cellcolor[rgb]{ .984,  .953,  .965}-0.87\% & 42.61\% & \cellcolor[rgb]{ .984,  .859,  .871}-3.27\% & 67.27\% & \cellcolor[rgb]{ .961,  .969,  .992}0.68\% & 70.34\% & \cellcolor[rgb]{ .984,  .925,  .937}-1.54\% & 67.97\% & \cellcolor[rgb]{ .82,  .871,  .941}4.01\% & 33.59\% & \cellcolor[rgb]{ .984,  .984,  1}0.15\% & 18.12\% & \cellcolor[rgb]{ .894,  .922,  .969}2.30\% & 10.08\% & \cellcolor[rgb]{ .925,  .945,  .98}1.51\% \\
        w/o role prompting & 67.74\% & \cellcolor[rgb]{ .988,  .988,  1}0.00\% & 42.77\% & \cellcolor[rgb]{ .984,  .867,  .878}-3.11\% & 66.33\% & \cellcolor[rgb]{ .984,  .976,  .988}-0.26\% & 70.19\% & \cellcolor[rgb]{ .984,  .922,  .933}-1.69\% & 58.63\% & \cellcolor[rgb]{ .98,  .78,  .792}-5.33\% & 25.31\% & \cellcolor[rgb]{ .976,  .675,  .682}-8.13\% & 9.94\% & \cellcolor[rgb]{ .98,  .761,  .769}-5.88\% & 5.40\% & \cellcolor[rgb]{ .984,  .863,  .875}-3.17\% \\
    \midrule
    \textbf{Storing Lan. JSON} & 68.04\% & \cellcolor[rgb]{ .988,  .988,  1}0.00\% & 42.40\% & \cellcolor[rgb]{ .988,  .988,  1}0.00\% & 70.39\% & \cellcolor[rgb]{ .988,  .988,  1}0.00\% & 73.84\% & \cellcolor[rgb]{ .988,  .988,  1}0.00\% & 63.99\% & \cellcolor[rgb]{ .988,  .988,  1}0.00\% & 28.14\% & \cellcolor[rgb]{ .988,  .988,  1}0.00\% & 15.23\% & \cellcolor[rgb]{ .988,  .988,  1}0.00\% & 8.82\% & \cellcolor[rgb]{ .988,  .988,  1}0.00\% \\
        w/o format explanation & 67.96\% & \cellcolor[rgb]{ .984,  .984,  .996}-0.08\% & 41.35\% & \cellcolor[rgb]{ .984,  .945,  .957}-1.05\% & 70.85\% & \cellcolor[rgb]{ .949,  .961,  .988}0.98\% & 73.51\% & \cellcolor[rgb]{ .984,  .973,  .984}-0.33\% & 72.59\% & \cellcolor[rgb]{ .627,  .733,  .875}8.60\% & 31.78\% & \cellcolor[rgb]{ .835,  .882,  .949}3.63\% & 20.30\% & \cellcolor[rgb]{ .776,  .839,  .925}5.07\% & 11.73\% & \cellcolor[rgb]{ .867,  .902,  .957}2.91\% \\
        w/o partition mark & 67.65\% & \cellcolor[rgb]{ .984,  .973,  .984}-0.40\% & 42.40\% & \cellcolor[rgb]{ .988,  .988,  1}0.00\% & 69.96\% & \cellcolor[rgb]{ .988,  .988,  1}0.09\% & 72.83\% & \cellcolor[rgb]{ .984,  .949,  .961}-1.01\% & 63.97\% & \cellcolor[rgb]{ .984,  .984,  .996}-0.02\% & 33.97\% & \cellcolor[rgb]{ .745,  .816,  .914}5.83\% & 17.54\% & \cellcolor[rgb]{ .894,  .922,  .969}2.30\% & 10.14\% & \cellcolor[rgb]{ .933,  .949,  .98}1.32\% \\
        w/o role prompting & 67.60\% & \cellcolor[rgb]{ .988,  .988,  1}0.00\% & 42.15\% & \cellcolor[rgb]{ .984,  .976,  .988}-0.25\% & 69.87\% & \cellcolor[rgb]{ .969,  .976,  .996}0.53\% & 72.99\% & \cellcolor[rgb]{ .984,  .953,  .965}-0.85\% & 50.39\% & \cellcolor[rgb]{ .973,  .463,  .471}-13.59\% & 24.65\% & \cellcolor[rgb]{ .984,  .851,  .863}-3.50\% & 12.75\% & \cellcolor[rgb]{ .984,  .89,  .902}-2.48\% & 7.13\% & \cellcolor[rgb]{ .984,  .922,  .933}-1.70\% \\
    \midrule
    \textbf{Markup Lan. XML} & 70.00\% & \cellcolor[rgb]{ .988,  .988,  1}0.00\% & \textbf{47.20\%} & \cellcolor[rgb]{ .988,  .988,  1}0.00\% & 70.74\% & \cellcolor[rgb]{ .988,  .988,  1}0.00\% & 73.14\% & \cellcolor[rgb]{ .988,  .988,  1}0.00\% & 52.26\% & \cellcolor[rgb]{ .988,  .988,  1}0.00\% & 32.16\% & \cellcolor[rgb]{ .988,  .988,  1}0.00\% & 15.23\% & \cellcolor[rgb]{ .988,  .988,  1}0.00\% & 8.82\% & \cellcolor[rgb]{ .988,  .988,  1}0.00\% \\
        w/o format explanation & \textbf{70.45\%} & \cellcolor[rgb]{ .973,  .976,  .996}0.45\% & 43.92\% & \cellcolor[rgb]{ .984,  .859,  .871}-3.28\% & \textbf{71.74\%} & \cellcolor[rgb]{ .949,  .961,  .988}1.01\% & 72.03\% & \cellcolor[rgb]{ .984,  .945,  .957}-1.10\% & 58.66\% & \cellcolor[rgb]{ .718,  .8,  .906}6.40\% & 36.57\% & \cellcolor[rgb]{ .804,  .859,  .937}4.42\% & 17.12\% & \cellcolor[rgb]{ .91,  .933,  .973}1.88\% & 10.06\% & \cellcolor[rgb]{ .937,  .953,  .984}1.23\% \\
        w/o partition mark & 70.16\% & \cellcolor[rgb]{ .984,  .984,  1}0.15\% & 44.01\% & \cellcolor[rgb]{ .984,  .863,  .875}-3.19\% & 70.77\% & \cellcolor[rgb]{ .988,  .988,  1}0.03\% & 71.18\% & \cellcolor[rgb]{ .984,  .91,  .922}-1.96\% & 62.77\% & \cellcolor[rgb]{ .545,  .678,  .847}10.51\% & 35.40\% & \cellcolor[rgb]{ .851,  .894,  .953}3.24\% & 18.69\% & \cellcolor[rgb]{ .843,  .886,  .949}3.45\% & 10.97\% & \cellcolor[rgb]{ .898,  .925,  .969}2.14\% \\
        w/o role prompting & 70.30\% & \cellcolor[rgb]{ .976,  .98,  .996}0.30\% & 44.65\% & \cellcolor[rgb]{ .984,  .886,  .898}-2.55\% & 69.47\% & \cellcolor[rgb]{ .984,  .937,  .949}-1.26\% & 72.43\% & \cellcolor[rgb]{ .984,  .961,  .973}-0.71\% & 43.45\% & \cellcolor[rgb]{ .976,  .647,  .659}-8.81\% & 20.32\% & \cellcolor[rgb]{ .973,  .529,  .541}-11.84\% & 14.05\% & \cellcolor[rgb]{ .984,  .941,  .953}-1.18\% & 8.23\% & \cellcolor[rgb]{ .984,  .965,  .976}-0.60\% \\
    \midrule
    \textbf{Markup Lan. HTML} & \textbf{71.33\%} & \cellcolor[rgb]{ .988,  .988,  1}0.00\% & \textbf{47.29\%} & \cellcolor[rgb]{ .988,  .988,  1}0.00\% & 71.31\% & \cellcolor[rgb]{ .988,  .988,  1}0.00\% & \textbf{75.20\%} & \cellcolor[rgb]{ .988,  .988,  1}0.00\% & \textbf{79.04\%} & \cellcolor[rgb]{ .988,  .988,  1}0.00\% & 39.72\% & \cellcolor[rgb]{ .988,  .988,  1}0.00\% & 21.45\% & \cellcolor[rgb]{ .988,  .988,  1}0.00\% & 12.30\% & \cellcolor[rgb]{ .988,  .988,  1}0.00\% \\
        w/o format explanation & 70.28\% & \cellcolor[rgb]{ .984,  .945,  .957}-1.04\% & 44.00\% & \cellcolor[rgb]{ .984,  .859,  .871}-3.29\% & \textbf{72.19\%} & \cellcolor[rgb]{ .953,  .965,  .988}0.88\% & 67.92\% & \cellcolor[rgb]{ .98,  .706,  .718}-7.28\% & \textbf{73.45\%} & \cellcolor[rgb]{ .98,  .773,  .78}-5.60\% & \textbf{43.00\%} & \cellcolor[rgb]{ .851,  .894,  .953}3.27\% & \textbf{23.31\%} & \cellcolor[rgb]{ .91,  .933,  .973}1.86\% & \textbf{14.53\%} & \cellcolor[rgb]{ .894,  .925,  .969}2.22\% \\
        w/o partition mark & 69.93\% & \cellcolor[rgb]{ .984,  .933,  .945}-1.40\% & 44.83\% & \cellcolor[rgb]{ .984,  .89,  .902}-2.45\% & 71.59\% & \cellcolor[rgb]{ .976,  .98,  .996}0.28\% & 71.05\% & \cellcolor[rgb]{ .98,  .827,  .839}-4.15\% & \textbf{76.35\%} & \cellcolor[rgb]{ .98,  .773,  .78}-5.60\% & 40.84\% & \cellcolor[rgb]{ .941,  .957,  .984}1.11\% & 21.75\% & \cellcolor[rgb]{ .976,  .98,  .996}0.30\% & 12.51\% & \cellcolor[rgb]{ .98,  .984,  1}0.21\% \\
        w/o role prompting & \textbf{70.30\%} & \cellcolor[rgb]{ .984,  .945,  .957}-1.03\% & 44.29\% & \cellcolor[rgb]{ .984,  .871,  .882}-2.99\% & 70.48\% & \cellcolor[rgb]{ .984,  .922,  .933}-1.71\% & 65.00\% & \cellcolor[rgb]{ .976,  .596,  .604}-10.20\% & 69.80\% & \cellcolor[rgb]{ .976,  .631,  .639}-9.25\% & 31.43\% & \cellcolor[rgb]{ .976,  .667,  .678}-8.29\% & 17.15\% & \cellcolor[rgb]{ .98,  .82,  .831}-4.30\% & 9.74\% & \cellcolor[rgb]{ .984,  .886,  .898}-2.56\% \\
    \bottomrule
    \end{tabular}}
\end{table*}
\begin{table*}[htbp]
  \centering
  \caption{Full results of the benchmark.}
  \label{tab:benchmark}
  \resizebox{\textwidth}{!}{
    \begin{tabular}{lcccccccccccccc}
    \toprule
    \multicolumn{1}{c}{\multirow{2}[4]{*}{Input Design}} & \multicolumn{2}{c}{Table Partition} & \multicolumn{2}{c}{Cell Lookup} & \multicolumn{2}{c}{Reverse Lookup} & \multicolumn{2}{c}{Column Retrieval} & \multicolumn{2}{c}{Row Retrieval} & \multicolumn{2}{c}{Size Detection} & \multicolumn{2}{c}{Merged Cell Detection} \\
\cmidrule{2-15}          & Acc   & $\Delta$     & Acc   & $\Delta$     & Acc   & $\Delta$     & Acc   & $\Delta$     & Acc   & $\Delta$     & Acc   & $\Delta$     & Acc   & $\Delta$ \\
    \midrule
    \textbf{NL + Sep} & 93.00\% & \cellcolor[rgb]{ .988,  .988,  1}0.00\% & 39.67\% & \cellcolor[rgb]{ .988,  .988,  1}0.00\% & 52.00\% & \cellcolor[rgb]{ .988,  .988,  1}0.00\% & 60.67\% & \cellcolor[rgb]{ .988,  .988,  1}0.00\% & 31.00\% & \cellcolor[rgb]{ .988,  .988,  1}0.00\% & 42.00\% & \cellcolor[rgb]{ .988,  .988,  1}0.00\% & 71.33\% & \cellcolor[rgb]{ .988,  .988,  1}0.00\% \\
        w/o format explanation & 91.33\% & \cellcolor[rgb]{ .984,  .922,  .933}-1.67\% & 50.00\% & \cellcolor[rgb]{ .553,  .682,  .847}10.33\% & \textbf{58.33\%} & \cellcolor[rgb]{ .722,  .8,  .906}6.33\% & 58.00\% & \cellcolor[rgb]{ .984,  .882,  .894}-2.67\% & 31.67\% & \cellcolor[rgb]{ .961,  .969,  .992}0.67\% & 40.67\% & \cellcolor[rgb]{ .984,  .933,  .945}-1.33\% & 73.33\% & \cellcolor[rgb]{ .906,  .929,  .973}2.00\% \\
        w/o partition mark & 90.00\% & \cellcolor[rgb]{ .984,  .871,  .882}-3.00\% & 40.33\% & \cellcolor[rgb]{ .961,  .969,  .992}0.67\% & 50.00\% & \cellcolor[rgb]{ .984,  .91,  .922}-2.00\% & 56.67\% & \cellcolor[rgb]{ .98,  .831,  .843}-4.00\% & 36.33\% & \cellcolor[rgb]{ .765,  .831,  .922}5.33\% & 40.00\% & \cellcolor[rgb]{ .984,  .91,  .922}-2.00\% & 69.00\% & \cellcolor[rgb]{ .984,  .898,  .906}-2.33\% \\
        w/o role prompting & 91.33\% & \cellcolor[rgb]{ .984,  .922,  .933}-1.67\% & 45.67\% & \cellcolor[rgb]{ .737,  .812,  .914}6.00\% & 41.00\% & \cellcolor[rgb]{ .976,  .565,  .573}-11.00\% & 54.00\% & \cellcolor[rgb]{ .98,  .729,  .741}-6.67\% & 25.33\% & \cellcolor[rgb]{ .98,  .769,  .78}-5.67\% & 41.67\% & \cellcolor[rgb]{ .984,  .973,  .984}-0.33\% & 74.00\% & \cellcolor[rgb]{ .878,  .91,  .961}2.67\% \\
        w/o change order & 88.00\% & \cellcolor[rgb]{ .98,  .792,  .804}-5.00\% & 35.33\% & \cellcolor[rgb]{ .98,  .82,  .831}-4.33\% & 40.33\% & \cellcolor[rgb]{ .973,  .537,  .545}-11.67\% & 55.33\% & \cellcolor[rgb]{ .98,  .78,  .792}-5.33\% & 24.67\% & \cellcolor[rgb]{ .98,  .741,  .753}-6.33\% & 36.67\% & \cellcolor[rgb]{ .98,  .78,  .792}-5.33\% & 61.00\% & \cellcolor[rgb]{ .976,  .588,  .6}-10.33\% \\
    \midrule
    \textbf{    w/o 1-shot} & 75.33\% & \cellcolor[rgb]{ .973,  .412,  .42}-17.67\% & 9.00\% & \cellcolor[rgb]{ .973,  .412,  .42}-30.67\% & 37.33\% & \cellcolor[rgb]{ .973,  .424,  .431}-14.67\% & 50.33\% & \cellcolor[rgb]{ .976,  .588,  .6}-10.33\% & 17.00\% & \cellcolor[rgb]{ .973,  .447,  .455}-14.00\% & 26.33\% & \cellcolor[rgb]{ .973,  .412,  .42}-15.67\% & 17.33\% & \cellcolor[rgb]{ .973,  .412,  .42}-54.00\% \\
    \midrule
    \textbf{Storing Lang. JSON} & 94.00\% & \cellcolor[rgb]{ .988,  .988,  1}0.00\% & 42.67\% & \cellcolor[rgb]{ .988,  .988,  1}0.00\% & 54.33\% & \cellcolor[rgb]{ .988,  .988,  1}0.00\% & 54.33\% & \cellcolor[rgb]{ .988,  .988,  1}0.00\% & 29.00\% & \cellcolor[rgb]{ .988,  .988,  1}0.00\% & 42.67\% & \cellcolor[rgb]{ .988,  .988,  1}0.00\% & 73.33\% & \cellcolor[rgb]{ .988,  .988,  1}0.00\% \\
        w/o format explanation & 87.67\% & \cellcolor[rgb]{ .98,  .741,  .753}-6.33\% & 47.67\% & \cellcolor[rgb]{ .776,  .839,  .925}5.00\% & 57.67\% & \cellcolor[rgb]{ .851,  .89,  .953}3.33\% & 49.00\% & \cellcolor[rgb]{ .98,  .78,  .792}-5.33\% & 30.33\% & \cellcolor[rgb]{ .933,  .949,  .98}1.33\% & 39.67\% & \cellcolor[rgb]{ .984,  .871,  .882}-3.00\% & 71.67\% & \cellcolor[rgb]{ .984,  .922,  .933}-1.67\% \\
        w/o partition mark & 92.00\% & \cellcolor[rgb]{ .984,  .91,  .922}-2.00\% & 48.67\% & \cellcolor[rgb]{ .737,  .812,  .914}6.00\% & 44.00\% & \cellcolor[rgb]{ .976,  .588,  .6}-10.33\% & 59.67\% & \cellcolor[rgb]{ .765,  .831,  .922}5.33\% & 40.33\% & \cellcolor[rgb]{ .51,  .651,  .831}11.33\% & 39.67\% & \cellcolor[rgb]{ .984,  .871,  .882}-3.00\% & 72.67\% & \cellcolor[rgb]{ .984,  .961,  .973}-0.67\% \\
        w/o role prompting & 90.67\% & \cellcolor[rgb]{ .984,  .859,  .871}-3.33\% & 44.67\% & \cellcolor[rgb]{ .906,  .929,  .973}2.00\% & 41.67\% & \cellcolor[rgb]{ .973,  .498,  .51}-12.67\% & 57.33\% & \cellcolor[rgb]{ .863,  .902,  .957}3.00\% & 29.33\% & \cellcolor[rgb]{ .976,  .98,  .996}0.33\% & 38.33\% & \cellcolor[rgb]{ .98,  .82,  .831}-4.33\% & 72.67\% & \cellcolor[rgb]{ .984,  .961,  .973}-0.67\% \\
        w/o change order & 90.67\% & \cellcolor[rgb]{ .984,  .859,  .871}-3.33\% & 37.67\% & \cellcolor[rgb]{ .98,  .796,  .804}-5.00\% & 42.67\% & \cellcolor[rgb]{ .973,  .537,  .545}-11.67\% & 55.67\% & \cellcolor[rgb]{ .933,  .949,  .98}1.33\% & 26.00\% & \cellcolor[rgb]{ .984,  .871,  .882}-3.00\% & 33.00\% & \cellcolor[rgb]{ .976,  .616,  .624}-9.67\% & 62.67\% & \cellcolor[rgb]{ .976,  .576,  .584}-10.67\% \\
    \midrule
    \textbf{    w/o 1-shot} & 74.67\% & \cellcolor[rgb]{ .973,  .412,  .42}-19.33\% & 11.33\% & \cellcolor[rgb]{ .973,  .412,  .42}-31.33\% & 30.00\% & \cellcolor[rgb]{ .973,  .412,  .42}-24.33\% & 52.67\% & \cellcolor[rgb]{ .984,  .922,  .933}-1.67\% & 14.00\% & \cellcolor[rgb]{ .973,  .412,  .42}-15.00\% & 27.33\% & \cellcolor[rgb]{ .973,  .412,  .42}-15.33\% & 19.67\% & \cellcolor[rgb]{ .973,  .412,  .42}-53.67\% \\
    \midrule
    \textbf{Markup Lang. Markdown} & 92.33\% & \cellcolor[rgb]{ .988,  .988,  1}0.00\% & 43.33\% & \cellcolor[rgb]{ .988,  .988,  1}0.00\% & 51.00\% & \cellcolor[rgb]{ .988,  .988,  1}0.00\% & 35.33\% & \cellcolor[rgb]{ .988,  .988,  1}0.00\% & \textbf{42.33\%} & \cellcolor[rgb]{ .988,  .988,  1}0.00\% & 40.67\% & \cellcolor[rgb]{ .988,  .988,  1}0.00\% & \textbf{78.00\%} & \cellcolor[rgb]{ .988,  .988,  1}0.00\% \\
        w/o format explanation & 88.00\% & \cellcolor[rgb]{ .98,  .82,  .831}-4.33\% & \textbf{56.00\%} & \cellcolor[rgb]{ .455,  .612,  .812}12.67\% & 50.67\% & \cellcolor[rgb]{ .984,  .973,  .984}-0.33\% & 34.33\% & \cellcolor[rgb]{ .984,  .949,  .961}-1.00\% & 33.33\% & \cellcolor[rgb]{ .976,  .639,  .651}-9.00\% & 39.00\% & \cellcolor[rgb]{ .984,  .922,  .933}-1.67\% & 74.00\% & \cellcolor[rgb]{ .98,  .831,  .843}-4.00\% \\
        w/o partition mark & 96.33\% & \cellcolor[rgb]{ .82,  .871,  .941}4.00\% & 52.67\% & \cellcolor[rgb]{ .596,  .714,  .863}9.33\% & 54.67\% & \cellcolor[rgb]{ .835,  .882,  .949}3.67\% & 35.33\% & \cellcolor[rgb]{ .988,  .988,  1}0.00\% & \textbf{45.00\%} & \cellcolor[rgb]{ .878,  .91,  .961}2.67\% & 43.00\% & \cellcolor[rgb]{ .89,  .922,  .969}2.33\% & 74.67\% & \cellcolor[rgb]{ .984,  .859,  .871}-3.33\% \\
        w/o role prompting & 94.33\% & \cellcolor[rgb]{ .906,  .929,  .973}2.00\% & 48.33\% & \cellcolor[rgb]{ .78,  .843,  .929}5.00\% & \textbf{58.67\%} & \cellcolor[rgb]{ .667,  .761,  .886}7.67\% & 35.67\% & \cellcolor[rgb]{ .976,  .98,  .996}0.33\% & 39.67\% & \cellcolor[rgb]{ .984,  .882,  .894}-2.67\% & 40.00\% & \cellcolor[rgb]{ .984,  .961,  .973}-0.67\% & \textbf{78.00\%} & \cellcolor[rgb]{ .988,  .988,  1}0.00\% \\
        w/o change order & 89.67\% & \cellcolor[rgb]{ .984,  .882,  .894}-2.67\% & 40.33\% & \cellcolor[rgb]{ .984,  .871,  .882}-3.00\% & 52.00\% & \cellcolor[rgb]{ .949,  .961,  .988}1.00\% & 36.67\% & \cellcolor[rgb]{ .933,  .949,  .98}1.33\% & 34.00\% & \cellcolor[rgb]{ .976,  .667,  .675}-8.33\% & 24.67\% & \cellcolor[rgb]{ .973,  .412,  .42}-16.00\% & 59.67\% & \cellcolor[rgb]{ .973,  .412,  .42}-18.33\% \\
    \midrule
    \textbf{    w/o 1-shot} & 60.70\% & \cellcolor[rgb]{ .973,  .412,  .42}-31.64\% & 8.67\% & \cellcolor[rgb]{ .973,  .412,  .42}-34.67\% & 35.33\% & \cellcolor[rgb]{ .973,  .412,  .42}-15.67\% & 30.67\% & \cellcolor[rgb]{ .98,  .808,  .816}-4.67\% & 19.00\% & \cellcolor[rgb]{ .973,  .412,  .42}-23.33\% & 11.67\% & \cellcolor[rgb]{ .973,  .412,  .42}-29.00\% & 23.67\% & \cellcolor[rgb]{ .973,  .412,  .42}-54.33\% \\
    \midrule
    \textbf{Markup Lang. XML} & 96.00\% & \cellcolor[rgb]{ .988,  .988,  1}0.00\% & 43.33\% & \cellcolor[rgb]{ .988,  .988,  1}0.00\% & 55.00\% & \cellcolor[rgb]{ .988,  .988,  1}0.00\% & 41.33\% & \cellcolor[rgb]{ .988,  .988,  1}0.00\% & 41.00\% & \cellcolor[rgb]{ .988,  .988,  1}0.00\% & 43.67\% & \cellcolor[rgb]{ .988,  .988,  1}0.00\% & 75.00\% & \cellcolor[rgb]{ .988,  .988,  1}0.00\% \\
        w/o format explanation & 89.00\% & \cellcolor[rgb]{ .98,  .718,  .725}-7.00\% & \textbf{58.33\%} & \cellcolor[rgb]{ .353,  .541,  .776}15.00\% & 51.33\% & \cellcolor[rgb]{ .984,  .847,  .855}-3.67\% & 35.33\% & \cellcolor[rgb]{ .98,  .757,  .765}-6.00\% & 32.67\% & \cellcolor[rgb]{ .976,  .667,  .675}-8.33\% & 37.67\% & \cellcolor[rgb]{ .98,  .757,  .765}-6.00\% & 74.00\% & \cellcolor[rgb]{ .984,  .949,  .961}-1.00\% \\
        w/o partition mark & 96.33\% & \cellcolor[rgb]{ .976,  .98,  .996}0.33\% & 54.67\% & \cellcolor[rgb]{ .51,  .651,  .831}11.33\% & 55.00\% & \cellcolor[rgb]{ .988,  .988,  1}0.00\% & 36.00\% & \cellcolor[rgb]{ .98,  .78,  .792}-5.33\% & \textbf{48.00\%} & \cellcolor[rgb]{ .694,  .78,  .898}7.00\% & 39.33\% & \cellcolor[rgb]{ .98,  .82,  .831}-4.33\% & 74.33\% & \cellcolor[rgb]{ .984,  .961,  .973}-0.67\% \\
        w/o role prompting & 93.67\% & \cellcolor[rgb]{ .984,  .898,  .906}-2.33\% & 47.33\% & \cellcolor[rgb]{ .82,  .871,  .941}4.00\% & \textbf{60.33\%} & \cellcolor[rgb]{ .765,  .831,  .922}5.33\% & 42.33\% & \cellcolor[rgb]{ .949,  .961,  .988}1.00\% & 37.00\% & \cellcolor[rgb]{ .98,  .831,  .843}-4.00\% & 40.67\% & \cellcolor[rgb]{ .984,  .871,  .882}-3.00\% & 73.33\% & \cellcolor[rgb]{ .984,  .922,  .933}-1.67\% \\
        w/o change order & 88.67\% & \cellcolor[rgb]{ .98,  .706,  .714}-7.33\% & 42.33\% & \cellcolor[rgb]{ .984,  .949,  .961}-1.00\% & 49.00\% & \cellcolor[rgb]{ .98,  .757,  .765}-6.00\% & 37.67\% & \cellcolor[rgb]{ .984,  .847,  .855}-3.67\% & 33.33\% & \cellcolor[rgb]{ .976,  .69,  .702}-7.67\% & 27.00\% & \cellcolor[rgb]{ .973,  .412,  .42}-16.67\% & 57.00\% & \cellcolor[rgb]{ .973,  .412,  .42}-18.00\% \\
    \midrule
    \textbf{    w/o 1-shot} & 69.33\% & \cellcolor[rgb]{ .973,  .412,  .42}-26.67\% & 9.00\% & \cellcolor[rgb]{ .973,  .412,  .42}-34.33\% & 33.00\% & \cellcolor[rgb]{ .973,  .412,  .42}-22.00\% & 25.33\% & \cellcolor[rgb]{ .973,  .412,  .42}-16.00\% & 15.33\% & \cellcolor[rgb]{ .973,  .412,  .42}-25.67\% & 12.67\% & \cellcolor[rgb]{ .973,  .412,  .42}-31.00\% & 22.33\% & \cellcolor[rgb]{ .973,  .412,  .42}-52.67\% \\
    \midrule
    \textbf{Markup Lang. HTML} & \textbf{96.67\%} & \cellcolor[rgb]{ .988,  .988,  1}0.00\% & 44.00\% & \cellcolor[rgb]{ .988,  .988,  1}0.00\% & 47.33\% & \cellcolor[rgb]{ .988,  .988,  1}0.00\% & \textbf{63.33\%} & \cellcolor[rgb]{ .988,  .988,  1}0.00\% & 42.00\% & \cellcolor[rgb]{ .988,  .988,  1}0.00\% & 67.00\% & \cellcolor[rgb]{ .988,  .988,  1}0.00\% & 76.67\% & \cellcolor[rgb]{ .988,  .988,  1}0.00\% \\
        w/o format explanation & 92.00\% & \cellcolor[rgb]{ .98,  .808,  .816}-4.67\% & 52.00\% & \cellcolor[rgb]{ .651,  .753,  .882}8.00\% & 52.33\% & \cellcolor[rgb]{ .78,  .843,  .929}5.00\% & \textbf{64.33\%} & \cellcolor[rgb]{ .949,  .961,  .988}1.00\% & 36.00\% & \cellcolor[rgb]{ .98,  .757,  .765}-6.00\% & \textbf{78.00\%} & \cellcolor[rgb]{ .525,  .663,  .839}11.00\% & \textbf{77.67\%} & \cellcolor[rgb]{ .949,  .961,  .988}1.00\% \\
        w/o partition mark & \textbf{98.00\%} & \cellcolor[rgb]{ .933,  .949,  .98}1.33\% & \textbf{59.00\%} & \cellcolor[rgb]{ .357,  .545,  .78}15.00\% & 53.00\% & \cellcolor[rgb]{ .749,  .82,  .918}5.67\% & \textbf{66.00\%} & \cellcolor[rgb]{ .878,  .91,  .961}2.67\% & 39.67\% & \cellcolor[rgb]{ .984,  .898,  .906}-2.33\% & \textbf{72.00\%} & \cellcolor[rgb]{ .78,  .843,  .929}5.00\% & 70.33\% & \cellcolor[rgb]{ .98,  .741,  .753}-6.33\% \\
        w/o role prompting & 95.00\% & \cellcolor[rgb]{ .863,  .902,  .957}3.00\% & 40.67\% & \cellcolor[rgb]{ .973,  .549,  .561}-11.33\% & 44.67\% & \cellcolor[rgb]{ .976,  .69,  .702}-7.67\% & 59.00\% & \cellcolor[rgb]{ .98,  .78,  .792}-5.33\% & 39.33\% & \cellcolor[rgb]{ .851,  .89,  .953}3.33\% & \textbf{69.00\%} & \cellcolor[rgb]{ .976,  .639,  .651}-9.00\% & 76.00\% & \cellcolor[rgb]{ .984,  .922,  .933}-1.67\% \\
        w/o change order & \textbf{96.67\%} & \cellcolor[rgb]{ .988,  .988,  1}0.00\% & 52.33\% & \cellcolor[rgb]{ .639,  .741,  .878}8.33\% & 40.67\% & \cellcolor[rgb]{ .98,  .729,  .741}-6.67\% & 55.67\% & \cellcolor[rgb]{ .976,  .69,  .702}-7.67\% & 31.67\% & \cellcolor[rgb]{ .976,  .588,  .6}-10.33\% & 52.67\% & \cellcolor[rgb]{ .973,  .435,  .443}-14.33\% & 65.67\% & \cellcolor[rgb]{ .976,  .565,  .573}-11.00\% \\
    \midrule
    \textbf{    w/o 1-shot} & 63.00\% & \cellcolor[rgb]{ .973,  .412,  .42}-33.67\% & 9.33\% & \cellcolor[rgb]{ .973,  .412,  .42}-34.67\% & 17.33\% & \cellcolor[rgb]{ .973,  .412,  .42}-30.00\% & 50.00\% & \cellcolor[rgb]{ .973,  .475,  .482}-13.33\% & 30.00\% & \cellcolor[rgb]{ .973,  .525,  .533}-12.00\% & 16.67\% & \cellcolor[rgb]{ .973,  .412,  .42}-50.33\% & 38.00\% & \cellcolor[rgb]{ .973,  .412,  .42}-38.67\% \\
    \bottomrule
    \end{tabular}}
\end{table*}%

\end{document}